\documentclass{article}

\PassOptionsToPackage{sort, numbers, compress}{natbib}

\usepackage[preprint]{neurips_2023}

\usepackage[utf8]{inputenc} % allow utf-8 input
\usepackage[T1]{fontenc}    % use 8-bit T1 fonts
\usepackage[pagebackref,breaklinks,colorlinks]{hyperref}       % hyperlinks
\usepackage{url}            % simple URL typesetting
\usepackage{booktabs}       % professional-quality tables
\usepackage{amsfonts}       % blackboard math symbols
\usepackage{nicefrac}       % compact symbols for 1/2, etc.
\usepackage{microtype}      % microtypography
\usepackage{graphicx}       % graphicx
\usepackage{xcolor}         % colors
\usepackage{cuted}
\usepackage{caption}
\usepackage{amsmath}
\usepackage{amssymb}
\usepackage{xspace}
\usepackage{multirow}
\usepackage{pifont}
\usepackage{diagbox}

\usepackage[capitalize]{cleveref}
\crefname{section}{Sec.}{Secs.}
\Crefname{section}{Section}{Sections}
\Crefname{table}{Table}{Tables}
\crefname{table}{Tab.}{Tabs.}

\newcommand{\cmark}{\ding{51}}
\newcommand{\xmark}{\ding{55}}

\makeatletter
\renewcommand{\paragraph}{%
  \@startsection{paragraph}{4}%
  {\z@}{0em}{-1em}%
  {\normalfont\normalsize\bfseries}%
}
\makeatother

\makeatletter
\DeclareRobustCommand\onedot{\futurelet\@let@token\@onedot}
\def\@onedot{\ifx\@let@token.\else.\null\fi\xspace}

\def\eg{\emph{e.g}\onedot} 
\def\ie{\emph{i.e}\onedot}

\makeatother

\makeatletter
\newcommand{\printfnsymbol}[1]{%
  \textsuperscript{\@fnsymbol{#1}}%
}
\makeatother

\title{Understanding Self-Supervised Features for Learning \\ Unsupervised Instance Segmentation}

\author{Paul Engstler\thanks{Equal contribution.} \quad \quad
Luke Melas-Kyriazi\printfnsymbol{1} \quad \quad
Christian Rupprecht \quad \quad
Iro Laina\\
University of Oxford\\
\texttt{\{paule,lukemk,chrisr,iro\}@robots.ox.ac.uk}\\
}

\begin{document}

\maketitle

\begin{abstract}

Self-supervised learning (SSL) can be used to solve complex visual tasks without human labels. 
Self-supervised representations encode useful semantic information about images, and as a result, they have already been used for tasks such as unsupervised semantic segmentation.
In this paper, we investigate self-supervised representations for instance segmentation without any manual annotations.
We find that the features of different SSL methods vary in their level of instance-awareness. In particular, DINO features, which are known to be excellent semantic descriptors, lack behind MAE features in their sensitivity for separating instances.

\end{abstract}
\section{Introduction}\label{s:intro}

Self-supervised visual representation learning has received significant attention over the past years.
The goal is to learn strong feature representations from unlabeled images, %\ie without manual annotations, 
which are transferable to downstream tasks. %  such as image classification or object detection.   
Most methods are currently designed for object-centric datasets like  ImageNet~\cite{ILSVRC15}, yet most downstream applications are scene-centric, containing multiple objects of interest.

We know that such representations can ``store'' several \textit{different} concepts even when they appear simultaneously in an image~\cite{laina2022measuring}.
It is thus reasonable to assume that they can be used to discover \emph{multiple} objects in an image, without supervision.
However, is it also possible to locate each concept spatially in the image? To answer this question, recent work has moved beyond concept discovery at the image level to pixel-level localization and segmentation of unsupervised concepts~\cite{LOST, melas2022deep, wang2022freesolo, shin2022unsupervised, amir2021deep}.  
Different from fine-tuning models for detection and segmentation tasks, this line of work exploits self-supervised features directly without using any annotations,
for example through clustering~\cite{melas2022deep,hamilton2022unsupervised,vangansbeke2020unsupervised} for semantic segmentation.

Instance segmentation, on the other hand, has received less attention. 
This task is particularly challenging because it requires recognizing and segmenting each individual object in an image, while semantic segmentation treats multiple objects of the same category as one.  
In the unsupervised setting, instance segmentation relies on acquiring a notion of objectness, spanning diverse scales and appearances, that is more difficult to achieve than single-object discovery, which amounts to the most salient region in the image. 
As a result, prior work on this task mines a small set of pseudo-masks per image and uses that to bootstrap instance segmentation~\cite{wang2022freesolo, van2022discovering, wang2023cut} or object detection~\cite{LOST} models.

In this work, we provide valuable insights with respect to the \emph{choice of self-supervised features} that can be used to obtain these mask proposals.
We find that the training objective in self-supervised models influences their ability to distinguish between objects, specifically different instances of the same semantic category. 
\section{Related Work}\label{s:related}
% We propose a method for both class-agnostic and class-aware instance segmentation without labels.
% We categorize other related work as follows. 

\paragraph{Self-Supervised Representation Learning (SSL).}

The goal of SSL is to learn view-invariant representations from unlabeled data, which are then transferred to downstream tasks via task-specific finetuning.
Following instance discrimination~\cite{dosovitskiy2014discriminative,wu2018unsupervised}, the majority of works focus on contrastive~\cite{hjelm2018learning,chen2020simple,kalantidis2020hard,he2019moco,henaff2020data,tian2020makes,bachman2019learning,chen2021empirical}, negative-free~\cite{bardes2022vicreg, grill2020byol,caron2021emerging,zbontar2021barlow,chen2021exploring}, and clustering-based~\cite{asano2019self,caron2020unsupervised,li2021prototypical,caron18deepcluster,ji2019invariant} learning.
Inspired by advances in NLP, masked image modeling~\cite{zhou2022image,xie2022simmim,he2022masked,wei2022masked,bao2022beit} has emerged as an alternative approach.
Meanwhile, the use of Vision Transformers (ViTs)~\cite{touvron2020deit,dosovitskiy2020image} 
has contributed significantly to the performance of self-supervised methods. 
Finally, there is a growing interest in learning dense~\cite{xie2020propagate,wang2020DenseCL,o2020unsupervised,selvaraju2021casting} or region-based representations~\cite{xie2021detco,xie2021unsupervised,yang2021instance,wei2021aligning,bar2022detreg,henaff2022object,yun2022patch,henaff2021efficient}, since they may be better suited for dense prediction tasks. 

\paragraph{Unsupervised Object Discovery.}
Unsupervised object discovery aims to localize objects in images by predicting bounding boxes. 
Earlier approaches were based on adversarial learning~\cite{chen_unsupervised_2019,bielski2019emergence,arandjelovic2019object,abdal2021labels4free}, while, more recently, several works~\cite{amir2021deep,melas2022deep,wang2022self,LOST,shin2022unsupervised,van2022discovering,wang2023cut} have explored the use of self-supervised features.
In particular, \cite{caron2021emerging} first showed that the self-attention of DINO-ViT could be used for foreground segmentation.
Follow-up works \cite{LOST,melas2022deep,wang2022self,van2022discovering} showed that it can also be used for salient object discovery, which in turn can be used to train unsupervised object detectors~\cite{wang2023cut,bielski2022move,LOST}. 

\paragraph{Unsupervised Semantic Segmentation.}
Recent work in unsupervised semantic segmentation can be split into two categories:
(1) methods that utilize pre-trained self-supervised models for initialization~\cite{vangansbeke2020unsupervised,cho2021picie,ziegler2022self,wen2022self,van2022discovering} and 
(2) methods that directly exploit off-the-shelf self-supervised representations (\eg, DINO~\cite{caron2021emerging}) to obtain and cluster pseudo-masks and train a segmentation network~\cite{melas2022deep,van2022discovering,hamilton2022unsupervised}.
These works demonstrate that self-supervised ViT features encode well-localized \emph{semantics}, but do not investigate whether these features can discriminate \emph{instances} of the same semantic class.
We aim to answer this question in \cref{s:ssl-feats}, finding a notable difference between models trained with discriminative (\eg, contrastive) and generative (\eg, autoencoding) objectives. 

\paragraph{Unsupervised Instance Segmentation.}

Unsupervised instance segmentation refers to the task of discovering and segmenting individual object instances in an image.
State-of-the-art methods typically bootstrap class-agnostic instance segmentation networks using coarse masks extracted from self-supervised feature extractors. 
FreeSOLO~\cite{wang2022freesolo} uses densely-learned features (DenseCL~\cite{wang2020DenseCL}), while Exemplar-FreeSOLO~\cite{Ishtiak_2023_CVPR} additionally uses a pool of exemplar images to extract objects. 
MaskDistill~\cite{van2022discovering} and CutLER~\cite{wang2023cut} leverage DINO features; \cite{van2022discovering} follows a single-object discovery technique, while \cite{wang2023cut} obtains multiple masks through repeated applications of the NCut algorithm~\cite{shi2000normalized}.

\section{Analysis of \emph{Instance-ness} in Self-Supervised Vision Transformers}\label{s:ssl-feats}

We investigate whether features from self-supervised transformers exhibit a general notion of \emph{instance-ness} that can be used to discover and discriminate between object instances rather than semantic categories.
Among SSL methods, DINO features have been frequently used for dense unsupervised tasks.
We note that different pre-training objectives may result in models with different properties and thus focus our investigation on models trained with a variety of objectives: contrastive (MoCo v3~\cite{chen2021empirical}), self-distillation (DINO~\cite{caron2021emerging}, MSN~\cite{assran2022masked}), image reconstruction (MAE~\cite{he2022masked}), as well as patch-level (SelfPatch~\cite{yun2022patch}) and dense (DenseCL~\cite{wang2020DenseCL}) pre-training.
All models use ViT-B/16 backbones, except for SelfPatch (only available for ViT-S/16) and DenseCL (ResNet-101~\cite{resnet}).

\begin{figure}
\centering
\includegraphics[width=0.8\linewidth]{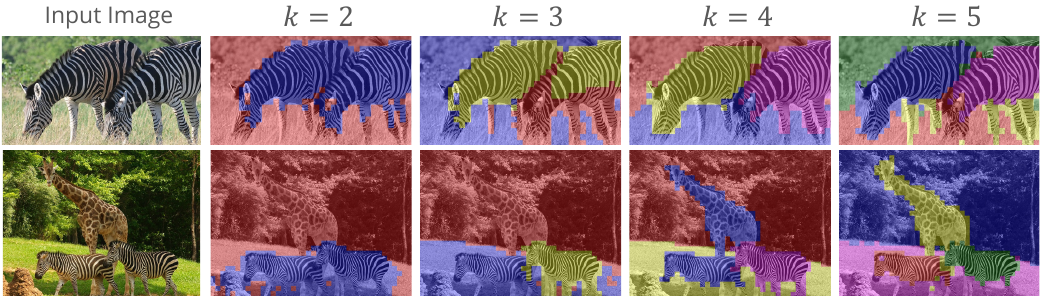}
\caption{\textbf{Varying the value of $k$.} Depending on the number of instances and semantic classes in an image, different values of $k$ capture scene elements at different levels of granularity. 
Higher values of $k$ separate the different instances. 
This example demonstrates why we use multiple values of $k$ to generate mask proposals and why it is necessary to filter these using a saliency map.
}
\label{fig:choice-of-k}
\end{figure}

First, we examine all models with respect to their ability to encode instances in the learned representation, without further task-specific finetuning.  
Similar to prior work~\cite{melas2022deep,amir2021deep}, to decompose an image into a set of meaningful regions, we compute an affinity matrix using cosine similarity in feature space and apply $k$-way spectral clustering spatially to generate a set of masks.
In particular, we extract features from the keys of the final self-attention block of ViTs.
To account for the varying number of objects in images and to allow for the possibility of segmenting objects at different granularities (\cref{fig:choice-of-k}), we apply spectral clustering multiple times for different values of $k \in K$ and accumulate all masks.
Finally, if the resulting masks have non-connected components, we split these off into separate masks.

It is already known from prior work that this process results in regions that likely correspond to semantic entities or objects.
To understand the degree to which these regions overlap with real objects in images, we compute the mean average recall of these masks against ground truth mask annotations in common datasets (MS COCO \texttt{val2017}~\cite{lin2014microsoft}, PASCAL VOC 2012~\cite{everingham2010pascal}) and report our findings in \Cref{tab:mini-experiment}.
We evaluate recall based on the number of instances of the same semantic category that co-occur in an image, \eg in an image with two objects of the same class and one object of another class, we compute and report recall for the two instances and the remaining single instance separately.

Interestingly, we find that MAE is much better at discriminating between multiple object instances (2+) than other models.
This finding may be explained by the fact that the pixel-wise reconstruction objective encourages the learning of representations of lower semanticity~\cite{he2022masked}.
As a result, it shows a lower tendency to group objects of the same or similar semantic category together, compared to, \eg, DINO. 
This holds true even against models trained with patch-level and pixel-level discrimination (\eg, SelfPatch). 
We demonstrate this also qualitatively in the Appendix (\Cref{tab:mini-experiment-images}), comparing MAE and DINO masks obtained after spectral clustering while varying $k$. 
Especially for higher values of $k$, MAE features tend to \emph{spatially} decompose an image, whereas DINO tends to divide objects further into semantic parts.
The spatial bias in MAE likely explains its superiority at separating instances, though MAE masks remain of generally lower quality than DINO.

At the same time, our results in \Cref{tab:mini-experiment} indicate that DINO and other contrastive methods such as MSN and MoCo-v3 are superior at locating \emph{single} objects; this finding matches prior work that uses DINO features for single-object image segmentation.

\definecolor{cadetgrey}{rgb}{0.57, 0.64, 0.69}
\newcommand{\arch}[1]{\textcolor{cadetgrey}{(#1)}}
\begin{table*}[t]
    \centering
    \small
    \caption{\textbf{Analysis of SSL methods for instance segmentation.} We report the mean average recall (in \%) of feature extractors for different numbers of instances of the same semantic class. For each image and feature extractor, we apply spectral clustering to the extracted features with $K = \{2,3,4,5,6\}$. 
    }\label{tab:mini-experiment}
    \def\arraystretch{0.95}
    \begin{tabular}{@{}lc@{\hskip 10pt}c@{\hskip 10pt}c@{\hskip 10pt}c@{\hskip 10pt}c@{\hskip 10pt}cc@{\hskip 10pt}c@{\hskip 10pt}c@{\hskip 10pt}c@{\hskip 10pt}c@{\hskip 10pt}c@{}}
        \toprule
          \multirow{2}{*}{\diagbox[trim=l,width=4cm,height=2.2\line]{\\{Features}}{Instances \\}}
           & \multicolumn{6}{c}{COCO} & \multicolumn{6}{c}{PASCAL VOC} \\
         \cmidrule(lr){2-7} \cmidrule(l){8-13}
         & 1 & 2 & 3 & 4 & 5 & 6+ & 1 & 2 & 3 & 4 & 5 & 6+ \\
         \midrule
         MAE \arch{ViT-B/16} \cite{he2022masked} & 16.8 & \textbf{8.6} & \textbf{6.3} & \textbf{3.2} & \textbf{2.9} & \textbf{1.1} & 31.7 & \textbf{18.1} & \textbf{9.3} & \textbf{4.5} & \textbf{4.1} & \textbf{2.2} \\
         DINO \arch{ViT-B/16} \cite{caron2021emerging} & \textbf{21.0} & 5.9 & 3.3 & 1.9 & 1.9 & 0.6 & \textbf{38.3} & 9.9 & 4.7 & 1.9 & 1.7 & 0.7 \\
         SelfPatch \arch{ViT-S/16} \cite{yun2022patch} & 17.4 & 7.7 & 5.5 & 2.8 & 2.6 & 1.0 & 32.8 & 13.9 & 7.1 & 4.3 & 3.4 & 1.5 \\
         MSN \arch{ViT-B/16} \cite{assran2022masked} & 19.8 & 5.7 & 3.4 & 1.9 & 1.8 & 0.6 & 35.6 & 10.1 & 4.9 & 1.7 & 1.3 & 0.7 \\
         MoCo-v3 \arch{ViT-B/16} \cite{chen2021empirical} & 19.1 & 5.0 & 3.0 & 1.6 & 1.7 & 0.6 & 37.3 & 9.1 & 4.3 & 1.6 & 1.6 & 0.7 \\
         DenseCL \arch{RN-101} \cite{wang2020DenseCL} & 13.3 & 3.2 & 1.8 & 0.9 & 0.9 & 0.3 & 24.3 & 5.2 & 2.3 & 1.0 & 0.8 & 0.3 \\
        \bottomrule
    \end{tabular}
\end{table*}

\section{Self-Training for Instance Segmentation}\label{s:method}

\begin{figure*}
  \centering
  \includegraphics[width=\textwidth]{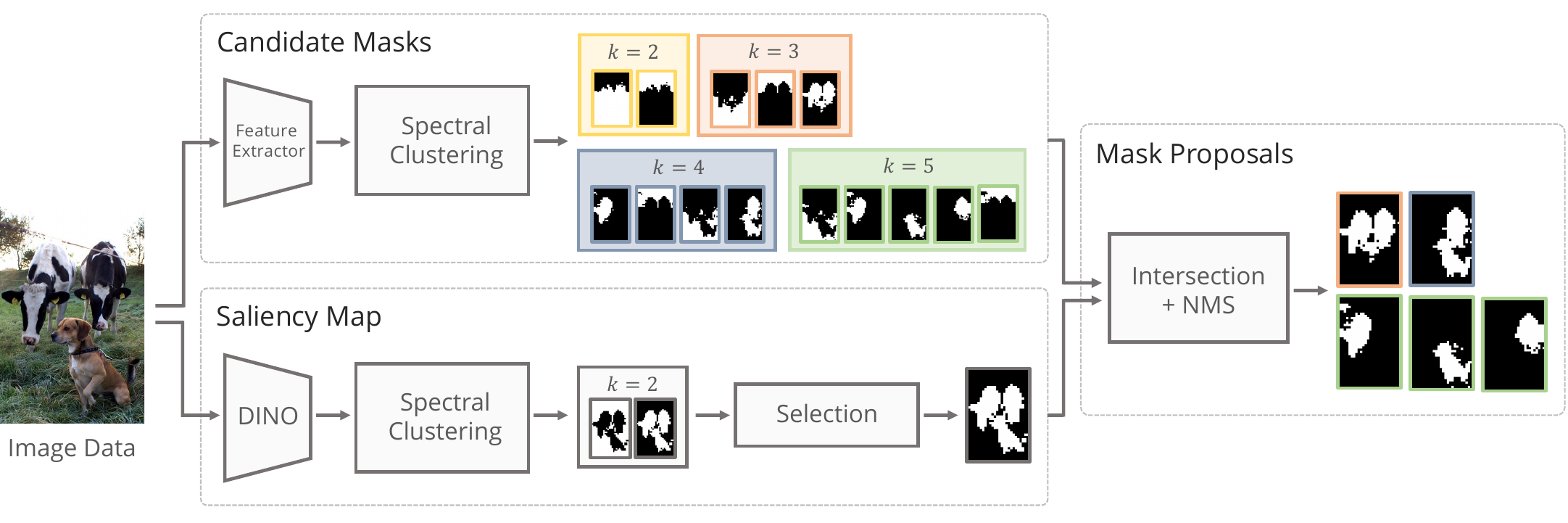}
  \caption{\textbf{Overview of our approach to produce mask proposals.} (1) Given an image and a self-supervised feature extractor, we generate a set of candidate masks by applying spectral clustering multiple times with different values for $k$, (2) we obtain a saliency map via spectral clustering with $k=2$ on DINO features, and (3) we select the candidate masks that strongly intersect with the saliency map as our final mask proposals (followed by non-maximum suppression for deduplication).}\label{fig:method}
\end{figure*}

To evaluate how these findings transfer to the task of unsupervised instance segmentation, we implement a simple method to generate mask proposals and use these to train a segmentation network, using the feature extractors considered above.
An overview of our approach is shown in \Cref{fig:method}.
 
We adopt a two-part approach to generate mask proposals, both parts based on SSL feature extractors. 
The first is the generation of pseudo-masks spanning the entire image area. 
For each of the feature extractors we follow the procedure outlined previously and apply multi-$k$-way clustering with $K=\{2,3,4,5\}$ (resulting in 14 overlapping masks per image). 
Since we cluster features spatially over the entire image area, the resulting clusters (masks) will inevitably also enclose background regions. 
As we are interested in \emph{instance} segmentation (\eg, people, cars) rather than \emph{stuff} segmentation (\eg, grass, sky), we need to eliminate masks that are less likely to correspond to objects.

Therefore, the second part is saliency-based masking to filter possibly noisy candidates, such as those corresponding to background regions.
Although not \emph{all} objects in a scene are necessarily salient, salient image regions are more likely to contain object instances than non-salient ones. 
Given the success of DINO features for this task~\cite{caron2021emerging, melas2022deep, LOST, wang2022self} and our findings in \Cref{tab:mini-experiment}, we apply spectral clustering with $k = 2$ on DINO features. 
We select the mask that shares fewer pixels with the image boundary as the foreground and use it to filter the mask candidates from the first stage such that only masks for salient objects remain.

Finally, we use the resulting mask proposals (after non-maximum suppression (NMS) for deduplication) to train an instance segmentation network. 
We train a SOLOv2~\cite{wang2020solov2} architecture using only pseudo-masks, following FreeSOLO~\cite{wang2022freesolo}. For more training details, please refer to \Cref{ss:app:details}.  

We train a segmenter for each of the feature extractors using their respective mask proposals.
This allows us to assess their performance in unsupervised instance segmentation and examine whether our initial observations align with post-training results.
We report average precision and recall on MS COCO \texttt{val2017} in \Cref{tab:mini-experiment-validation}.
We observe that the model trained with masks from MAE remains the best-performing one.
Interestingly, the DINO-based model significantly narrows the performance gap after training, whereas other feature extractors perform worse. 
We hypothesize that this comes from the fact that DINO masks are generally cleaner and capture object boundaries better, which is important for learning. This could also explain why the vast majority of the current state of the art (see \Cref{table:seg-all}) in unsupervised instance segmentation performs well, despite leveraging DINO features.

\begin{table}[t]
  \small
  \centering
  \def\arraystretch{0.95}
  \setlength{\tabcolsep}{4pt}
    \caption{\textbf{Unsupervised instance segmentation performance (COCO \texttt{val2017}).} We evaluate segmentation models trained with mask proposals from different feature extractors.
  } \label{tab:mini-experiment-validation}
  \begin{tabular}{lcccccc}
  \toprule
  Feature Extractor & AP$_\text{50}$ & AP$_\text{75}$ & AP & AR$_\text{1}$ & AR$_\text{10}$ & AR$_\text{100}$ \\
  \midrule
  MAE \cite{he2022masked} & \textbf{12.1} & \textbf{3.7} & \textbf{5.2} & \textbf{3.8} & 10.9 & \textbf{18.3} \\
  DINO \cite{caron2021emerging} & 11.8 & 3.6 & 5.0 & 3.7 & \textbf{11.4} & 18.0 \\
  SelfPatch \cite{yun2022patch} & 6.3 & 2.2 & 2.8 & 3.4 & 4.9 & 4.9 \\
  MSN \cite{assran2022masked} & 9.0 & 2.7 & 3.9 & 3.9 & 6.2 & 6.2 \\
  MoCo v3 \cite{chen2021empirical} & 7.2 & 2.5 & 3.3 & 3.4 & 6.4 & 9.6 \\
  DenseCL \cite{wang2020DenseCL} & 7.7 & 2.5 & 3.3 & 3.3 & 7.3 & 10.2 \\
  \bottomrule
  \end{tabular}
\end{table}
\section{Conclusion}\label{s:conclusions}
In this work, we study the properties of SSL methods for the task of instance segmentation. We find that different learning signals result in features with varying characteristics. DINO learns highly semantic features and is thus widely used for semantic tasks. We find that image reconstruction, \eg MAE, tasks are better suited to discriminate instances of the same class inside an image. This is an overlooked property that can potentially be used in many instance-specific downstream tasks.

\paragraph{Acknowledgements.} P.~E., C.~R., and I.~L. are supported by ERC-UNION- CoG-101001212. P.~E. is also supported by Meta Research and a fellowship from the German Academic Exchange Service (DAAD). L.~M.~K. is supported by the Rhodes Trust. C.~R. and I.~L. are also supported by VisualAI EP/T028572/1.

{\small\bibliographystyle{ieee_fullname}\bibliography{refs}}

\begin{thebibliography}{10}\itemsep=-1pt

\bibitem{abdal2021labels4free}
Rameen Abdal, Peihao Zhu, Niloy Mitra, and Peter Wonka.
\newblock Labels4free: Unsupervised segmentation using stylegan.
\newblock In {\em Proc. {ICCV}}, 2021.

\bibitem{amir2021deep}
Shir Amir, Yossi Gandelsman, Shai Bagon, and Tali Dekel.
\newblock Deep vit features as dense visual descriptors.
\newblock {\em arXiv preprint arXiv:2112.05814}, 2(3):4, 2021.

\bibitem{arandjelovic2019object}
Relja Arandjelovi{\'c} and Andrew Zisserman.
\newblock Object discovery with a copy-pasting gan.
\newblock {\em arXiv preprint arXiv:1905.11369}, 2019.

\bibitem{asano2019self}
YM Asano, C Rupprecht, and A Vedaldi.
\newblock Self-labelling via simultaneous clustering and representation learning.
\newblock In {\em International Conference on Learning Representations}, 2019.

\bibitem{assran2022masked}
Mahmoud Assran, Mathilde Caron, Ishan Misra, Piotr Bojanowski, Florian Bordes, Pascal Vincent, Armand Joulin, Mike Rabbat, and Nicolas Ballas.
\newblock Masked siamese networks for label-efficient learning.
\newblock In {\em European Conference on Computer Vision}, pages 456--473. Springer, 2022.

\bibitem{bachman2019learning}
Philip Bachman, R~Devon Hjelm, and William Buchwalter.
\newblock Learning representations by maximizing mutual information across views.
\newblock In {\em Advances in Neural Information Processing Systems}, pages 15509--15519, 2019.

\bibitem{bao2022beit}
Hangbo Bao, Li Dong, Songhao Piao, and Furu Wei.
\newblock {BE}it: {BERT} pre-training of image transformers.
\newblock In {\em International Conference on Learning Representations}, 2022.

\bibitem{bar2022detreg}
Amir Bar, Xin Wang, Vadim Kantorov, Colorado~J Reed, Roei Herzig, Gal Chechik, Anna Rohrbach, Trevor Darrell, and Amir Globerson.
\newblock Detreg: Unsupervised pretraining with region priors for object detection.
\newblock In {\em Proceedings of the IEEE/CVF Conference on Computer Vision and Pattern Recognition}, pages 14605--14615, 2022.

\bibitem{bardes2022vicreg}
Adrien Bardes, Jean Ponce, and Yann LeCun.
\newblock {VICR}eg: Variance-invariance-covariance regularization for self-supervised learning.
\newblock In {\em International Conference on Learning Representations}, 2022.

\bibitem{bielski2022move}
Adam Bielski and Paolo Favaro.
\newblock Move: Unsupervised movable object segmentation and detection.
\newblock {\em arXiv preprint arXiv:2210.07920}, 2022.

\bibitem{bielski2019emergence}
Adam~Jakub Bielski and Paolo Favaro.
\newblock Emergence of object segmentation in perturbed generative models.
\newblock {\em Advances in Neural Information Processing Systems (NIPS)}, 32, 2019.

\bibitem{caron18deepcluster}
M. Caron, P. Bojanowski, A. Joulin, and M. Douze.
\newblock Deep clustering for unsupervised learning of visual features.
\newblock In {\em Proc. {ECCV}}, 2018.

\bibitem{caron2020unsupervised}
Mathilde Caron, Ishan Misra, Julien Mairal, Priya Goyal, Piotr Bojanowski, and Armand Joulin.
\newblock Unsupervised learning of visual features by contrasting cluster assignments.
\newblock In {\em Proceedings of Advances in Neural Information Processing Systems (NeurIPS)}, 2020.

\bibitem{caron2021emerging}
Mathilde Caron, Hugo Touvron, Ishan Misra, Herv\'e J\'egou, Julien Mairal, Piotr Bojanowski, and Armand Joulin.
\newblock Emerging properties in self-supervised vision transformers.
\newblock In {\em Proceedings of the International Conference on Computer Vision (ICCV)}, 2021.

\bibitem{chen_unsupervised_2019}
Mickaël Chen, Thierry Artières, and Ludovic Denoyer.
\newblock Unsupervised {Object} {Segmentation} by {Redrawing}.
\newblock In {\em Proc. {NeurIPS}}, volume~32, 2019.

\bibitem{chen2020simple}
Ting Chen, Simon Kornblith, Mohammad Norouzi, and Geoffrey Hinton.
\newblock A simple framework for contrastive learning of visual representations.
\newblock In {\em International conference on machine learning}, pages 1597--1607. PMLR, 2020.

\bibitem{chen2021exploring}
Xinlei Chen and Kaiming He.
\newblock Exploring simple siamese representation learning.
\newblock In {\em Proceedings of the IEEE/CVF Conference on Computer Vision and Pattern Recognition}, pages 15750--15758, 2021.

\bibitem{chen2021empirical}
Xinlei Chen, Saining Xie, and Kaiming He.
\newblock An empirical study of training self-supervised vision transformers.
\newblock In {\em Proceedings of the IEEE/CVF International Conference on Computer Vision}, pages 9640--9649, 2021.

\bibitem{cho2021picie}
Jang~Hyun Cho, Utkarsh Mall, Kavita Bala, and Bharath Hariharan.
\newblock Picie: Unsupervised semantic segmentation using invariance and equivariance in clustering.
\newblock In {\em Proceedings of the IEEE/CVF Conference on Computer Vision and Pattern Recognition}, pages 16794--16804, 2021.

\bibitem{dosovitskiy2020image}
Alexey Dosovitskiy, Lucas Beyer, Alexander Kolesnikov, Dirk Weissenborn, Xiaohua Zhai, Thomas Unterthiner, Mostafa Dehghani, Matthias Minderer, Georg Heigold, Sylvain Gelly, et~al.
\newblock An image is worth 16x16 words: Transformers for image recognition at scale.
\newblock In {\em International Conference on Learning Representations}, 2020.

\bibitem{dosovitskiy2014discriminative}
Alexey Dosovitskiy, Jost~Tobias Springenberg, Martin Riedmiller, and Thomas Brox.
\newblock Discriminative unsupervised feature learning with convolutional neural networks.
\newblock In {\em Advances in neural information processing systems}, pages 766--774, 2014.

\bibitem{everingham2010pascal}
Mark Everingham, Luc Van~Gool, Christopher~KI Williams, John Winn, and Andrew Zisserman.
\newblock The pascal visual object classes (voc) challenge.
\newblock {\em International journal of computer vision}, 88(2):303--338, 2010.

\bibitem{grill2020byol}
Jean-Bastien Grill, Florian Strub, Florent Altch\'{e}, Corentin Tallec, Pierre Richemond, Elena Buchatskaya, Carl Doersch, Bernardo Avila~Pires, Zhaohan Guo, Mohammad Gheshlaghi~Azar, Bilal Piot, koray kavukcuoglu, Remi Munos, and Michal Valko.
\newblock Bootstrap your own latent - a new approach to self-supervised learning.
\newblock In H. Larochelle, M. Ranzato, R. Hadsell, M.~F. Balcan, and H. Lin, editors, {\em Advances in Neural Information Processing Systems}, volume~33, pages 21271--21284. Curran Associates, Inc., 2020.

\bibitem{hamilton2022unsupervised}
Mark Hamilton, Zhoutong Zhang, Bharath Hariharan, Noah Snavely, and William~T. Freeman.
\newblock Unsupervised semantic segmentation by distilling feature correspondences.
\newblock In {\em Proc. {ICLR}}, 2022.

\bibitem{he2022masked}
Kaiming He, Xinlei Chen, Saining Xie, Yanghao Li, Piotr Doll{\'a}r, and Ross Girshick.
\newblock Masked autoencoders are scalable vision learners.
\newblock In {\em Proceedings of the IEEE/CVF Conference on Computer Vision and Pattern Recognition}, pages 16000--16009, 2022.

\bibitem{he2021masked}
Kaiming He, Xinlei Chen, Saining Xie, Yanghao Li, Piotr Dollár, and Ross Girshick.
\newblock Masked autoencoders are scalable vision learners, 2021.

\bibitem{he2019moco}
Kaiming He, Haoqi Fan, Yuxin Wu, Saining Xie, and Ross Girshick.
\newblock Momentum contrast for unsupervised visual representation learning.
\newblock In {\em Conference on Computer Vision and Pattern Recognition}, 2019.

\bibitem{he2017mask}
Kaiming He, Georgia Gkioxari, Piotr Doll{\'a}r, and Ross Girshick.
\newblock Mask r-cnn.
\newblock In {\em Proceedings of the IEEE international conference on computer vision}, pages 2961--2969, 2017.

\bibitem{resnet}
Kaiming He, Xiangyu Zhang, Shaoqing Ren, and Jian Sun.
\newblock Deep residual learning for image recognition.
\newblock In {\em Proceedings of the IEEE conference on computer vision and pattern recognition}, pages 770--778, 2016.

\bibitem{henaff2020data}
Olivier Henaff.
\newblock Data-efficient image recognition with contrastive predictive coding.
\newblock In {\em International conference on machine learning}, pages 4182--4192. PMLR, 2020.

\bibitem{henaff2021efficient}
Olivier~J H{\'e}naff, Skanda Koppula, Jean-Baptiste Alayrac, Aaron Van~den Oord, Oriol Vinyals, and Jo{\~a}o Carreira.
\newblock Efficient visual pretraining with contrastive detection.
\newblock In {\em Proceedings of the IEEE/CVF International Conference on Computer Vision}, pages 10086--10096, 2021.

\bibitem{henaff2022object}
Olivier~J. H{\'e}naff, Skanda Koppula, Evan Shelhamer, Daniel Zoran, Andrew Jaegle, Andrew Zisserman, João Carreira, and Relja Arandjelović.
\newblock Object discovery and representation networks.
\newblock In {\em Proc. {ECCV}}, pages 123--143, 2022.

\bibitem{hjelm2018learning}
R~Devon Hjelm, Alex Fedorov, Samuel Lavoie-Marchildon, Karan Grewal, Phil Bachman, Adam Trischler, and Yoshua Bengio.
\newblock Learning deep representations by mutual information estimation and maximization.
\newblock In {\em International Conference on Learning Representations}, 2019.

\bibitem{Ishtiak_2023_CVPR}
Taoseef Ishtiak, Qing En, and Yuhong Guo.
\newblock Exemplar-freesolo: Enhancing unsupervised instance segmentation with exemplars.
\newblock In {\em Proceedings of the IEEE/CVF Conference on Computer Vision and Pattern Recognition (CVPR)}, pages 15424--15433, June 2023.

\bibitem{ji2019invariant}
Xu Ji, Joao~F Henriques, and Andrea Vedaldi.
\newblock Invariant information clustering for unsupervised image classification and segmentation.
\newblock In {\em Proceedings of the IEEE/CVF International Conference on Computer Vision}, pages 9865--9874, 2019.

\bibitem{kalantidis2020hard}
Yannis Kalantidis, Mert~Bulent Sariyildiz, Noe Pion, Philippe Weinzaepfel, and Diane Larlus.
\newblock Hard negative mixing for contrastive learning.
\newblock {\em Proc. {NeurIPS}}, 2020.

\bibitem{laina2022measuring}
Iro Laina, Yuki~M Asano, and Andrea Vedaldi.
\newblock Measuring the interpretability of unsupervised representations via quantized reversed probing.
\newblock In {\em Proc. {ICLR}}, 2022.

\bibitem{li2021prototypical}
Junnan Li, Pan Zhou, Caiming Xiong, and Steven Hoi.
\newblock Prototypical contrastive learning of unsupervised representations.
\newblock In {\em International Conference on Learning Representations}, 2021.

\bibitem{lin2014microsoft}
Tsung-Yi Lin, Michael Maire, Serge Belongie, James Hays, Pietro Perona, Deva Ramanan, Piotr Doll{\'a}r, and C~Lawrence Zitnick.
\newblock Microsoft coco: Common objects in context.
\newblock In {\em European conference on computer vision}, pages 740--755. Springer, 2014.

\bibitem{melas2022deep}
Luke Melas-Kyriazi, Christian Rupprecht, Iro Laina, and Andrea Vedaldi.
\newblock Deep spectral methods: A surprisingly strong baseline for unsupervised semantic segmentation and localization.
\newblock In {\em Proc. {CVPR}}, pages 8364--8375, 2022.

\bibitem{o2020unsupervised}
Pedro~O. Pinheiro, Amjad Almahairi, Ryan Benmalek, Florian Golemo, and Aaron~C Courville.
\newblock Unsupervised learning of dense visual representations.
\newblock {\em Advances in Neural Information Processing Systems}, 33:4489--4500, 2020.

\bibitem{ILSVRC15}
Olga Russakovsky, Jia Deng, Hao Su, Jonathan Krause, Sanjeev Satheesh, Sean Ma, Zhiheng Huang, Andrej Karpathy, Aditya Khosla, Michael Bernstein, Alexander~C. Berg, and Li Fei-Fei.
\newblock {ImageNet Large Scale Visual Recognition Challenge}.
\newblock {\em {IJCV}}, 115(3):211--252, 2015.

\bibitem{selvaraju2021casting}
Ramprasaath~R Selvaraju, Karan Desai, Justin Johnson, and Nikhil Naik.
\newblock Casting your model: Learning to localize improves self-supervised representations.
\newblock In {\em Proceedings of the IEEE/CVF Conference on Computer Vision and Pattern Recognition}, pages 11058--11067, 2021.

\bibitem{shi2000normalized}
Jianbo Shi and Jitendra Malik.
\newblock Normalized cuts and image segmentation.
\newblock {\em IEEE Transactions on pattern analysis and machine intelligence}, 22(8):888--905, 2000.

\bibitem{shin2022unsupervised}
Gyungin Shin, Samuel Albanie, and Weidi Xie.
\newblock Unsupervised salient object detection with spectral cluster voting.
\newblock In {\em Proceedings of the IEEE/CVF Conference on Computer Vision and Pattern Recognition}, pages 3971--3980, 2022.

\bibitem{LOST}
Oriane Sim\'eoni, Gilles Puy, Huy~V. Vo, Simon Roburin, Spyros Gidaris, Andrei Bursuc, Patrick P\'erez, Renaud Marlet, and Jean Ponce.
\newblock Localizing objects with self-supervised transformers and no labels.
\newblock In {\em Proc. {BMVC}}, November 2021.

\bibitem{tian2020makes}
Yonglong Tian, Chen Sun, Ben Poole, Dilip Krishnan, Cordelia Schmid, and Phillip Isola.
\newblock What makes for good views for contrastive learning?
\newblock {\em Advances in Neural Information Processing Systems}, 33:6827--6839, 2020.

\bibitem{tian2021boxinst}
Zhi Tian, Chunhua Shen, Xinlong Wang, and Hao Chen.
\newblock Boxinst: High-performance instance segmentation with box annotations.
\newblock In {\em Proceedings of the IEEE/CVF Conference on Computer Vision and Pattern Recognition}, pages 5443--5452, 2021.

\bibitem{touvron2020deit}
Hugo Touvron, Matthieu Cord, Matthijs Douze, Francisco Massa, Alexandre Sablayrolles, and Herv{\'e} J{\'e}gou.
\newblock Training data-efficient image transformers \& distillation through attention.
\newblock In {\em International Conference on Machine Learning}, pages 10347--10357. PMLR, 2021.

\bibitem{vangansbeke2020unsupervised}
Wouter Van~Gansbeke, Simon Vandenhende, Stamatios Georgoulis, and Luc Van~Gool.
\newblock Unsupervised semantic segmentation by contrasting object mask proposals.
\newblock In {\em International Conference on Computer Vision}, 2021.

\bibitem{van2022discovering}
Wouter Van~Gansbeke, Simon Vandenhende, and Luc Van~Gool.
\newblock Discovering object masks with transformers for unsupervised semantic segmentation.
\newblock {\em arXiv preprint arXiv:2206.06363}, 2022.

\bibitem{wang2021unidentified}
Weiyao Wang, Matt Feiszli, Heng Wang, and Du Tran.
\newblock Unidentified video objects: A benchmark for dense, open-world segmentation.
\newblock In {\em Proceedings of the IEEE/CVF International Conference on Computer Vision}, pages 10776--10785, 2021.

\bibitem{wang2023cut}
Xudong Wang, Rohit Girdhar, Stella~X Yu, and Ishan Misra.
\newblock Cut and learn for unsupervised object detection and instance segmentation.
\newblock In {\em Proceedings of the IEEE/CVF Conference on Computer Vision and Pattern Recognition}, pages 3124--3134, 2023.

\bibitem{wang2022freesolo}
Xinlong Wang, Zhiding Yu, Shalini De~Mello, Jan Kautz, Anima Anandkumar, Chunhua Shen, and Jose~M Alvarez.
\newblock Freesolo: Learning to segment objects without annotations.
\newblock In {\em Proc. {CVPR}}, pages 14176--14186, 2022.

\bibitem{wang2020solov2}
Xinlong Wang, Rufeng Zhang, Tao Kong, Lei Li, and Chunhua Shen.
\newblock Solov2: Dynamic and fast instance segmentation.
\newblock {\em Advances in Neural information processing systems}, 33:17721--17732, 2020.

\bibitem{wang2020DenseCL}
Xinlong Wang, Rufeng Zhang, Chunhua Shen, Tao Kong, and Lei Li.
\newblock Dense contrastive learning for self-supervised visual pre-training.
\newblock In {\em Proc. IEEE Conf. Computer Vision and Pattern Recognition (CVPR)}, 2021.

\bibitem{wang2022self}
Yangtao Wang, Xi Shen, Shell~Xu Hu, Yuan Yuan, James~L Crowley, and Dominique Vaufreydaz.
\newblock Self-supervised transformers for unsupervised object discovery using normalized cut.
\newblock In {\em Proceedings of the IEEE/CVF Conference on Computer Vision and Pattern Recognition}, pages 14543--14553, 2022.

\bibitem{wei2022masked}
Chen Wei, Haoqi Fan, Saining Xie, Chao-Yuan Wu, Alan Yuille, and Christoph Feichtenhofer.
\newblock Masked feature prediction for self-supervised visual pre-training.
\newblock In {\em Proceedings of the IEEE/CVF Conference on Computer Vision and Pattern Recognition}, pages 14668--14678, 2022.

\bibitem{wei2021aligning}
Fangyun Wei, Yue Gao, Zhirong Wu, Han Hu, and Stephen Lin.
\newblock Aligning pretraining for detection via object-level contrastive learning.
\newblock {\em Advances in Neural Information Processing Systems}, 34:22682--22694, 2021.

\bibitem{wen2022self}
Xin Wen, Bingchen Zhao, Anlin Zheng, Xiangyu Zhang, and Xiaojuan Qi.
\newblock Self-supervised visual representation learning with semantic grouping.
\newblock {\em arXiv preprint arXiv:2205.15288}, 2022.

\bibitem{wu2018unsupervised}
Zhirong Wu, Yuanjun Xiong, Stella~X Yu, and Dahua Lin.
\newblock Unsupervised feature learning via non-parametric instance discrimination.
\newblock In {\em Proceedings of the IEEE conference on computer vision and pattern recognition}, pages 3733--3742, 2018.

\bibitem{xie2021detco}
Enze Xie, Jian Ding, Wenhai Wang, Xiaohang Zhan, Hang Xu, Peize Sun, Zhenguo Li, and Ping Luo.
\newblock Detco: Unsupervised contrastive learning for object detection.
\newblock In {\em Proceedings of the IEEE/CVF International Conference on Computer Vision}, pages 8392--8401, 2021.

\bibitem{xie2021unsupervised}
Jiahao Xie, Xiaohang Zhan, Ziwei Liu, Yew~Soon Ong, and Chen~Change Loy.
\newblock Unsupervised object-level representation learning from scene images.
\newblock {\em Advances in Neural Information Processing Systems}, 34:28864--28876, 2021.

\bibitem{xie2020propagate}
Zhenda Xie, Yutong Lin, Zheng Zhang, Yue Cao, Stephen Lin, and Han Hu.
\newblock Propagate yourself: Exploring pixel-level consistency for unsupervised visual representation learning.
\newblock In {\em IEEE/CVF Conference on Computer Vision and Pattern Recognition (CVPR)}, 2021.

\bibitem{xie2022simmim}
Zhenda Xie, Zheng Zhang, Yue Cao, Yutong Lin, Jianmin Bao, Zhuliang Yao, Qi Dai, and Han Hu.
\newblock Simmim: A simple framework for masked image modeling.
\newblock In {\em Proceedings of the IEEE/CVF Conference on Computer Vision and Pattern Recognition}, pages 9653--9663, 2022.

\bibitem{yang2021instance}
Ceyuan Yang, Zhirong Wu, Bolei Zhou, and Stephen Lin.
\newblock Instance localization for self-supervised detection pretraining.
\newblock In {\em Proceedings of the IEEE/CVF Conference on Computer Vision and Pattern Recognition}, pages 3987--3996, 2021.

\bibitem{yun2022patch}
Sukmin Yun, Hankook Lee, Jaehyung Kim, and Jinwoo Shin.
\newblock Patch-level representation learning for self-supervised vision transformers.
\newblock In {\em Proceedings of the IEEE/CVF Conference on Computer Vision and Pattern Recognition}, pages 8354--8363, 2022.

\bibitem{zbontar2021barlow}
Jure Zbontar, Li Jing, Ishan Misra, Yann LeCun, and St{\'e}phane Deny.
\newblock Barlow twins: Self-supervised learning via redundancy reduction.
\newblock In {\em International Conference on Machine Learning}, pages 12310--12320. PMLR, 2021.

\bibitem{zhou2022image}
Jinghao Zhou, Chen Wei, Huiyu Wang, Wei Shen, Cihang Xie, Alan Yuille, and Tao Kong.
\newblock Image {BERT} pre-training with online tokenizer.
\newblock In {\em International Conference on Learning Representations}, 2022.

\bibitem{ziegler2022self}
Adrian Ziegler and Yuki~M Asano.
\newblock Self-supervised learning of object parts for semantic segmentation.
\newblock In {\em Proceedings of the IEEE/CVF Conference on Computer Vision and Pattern Recognition}, pages 14502--14511, 2022.

\end{thebibliography}

\clearpage
\appendix
\section{Appendix}

\subsection{Qualitative Analysis of Instance-ness in Self-Supervised Vision Transformers}

In \Cref{s:ssl-feats}, we have measured the mean average recall of masks produced by clustering features from different self-supervised networks for a different number of instances of the same semantic class on MS COCO \texttt{val2017} and PASCAL VOC 2012. 
We used this experiment to better understand the properties of the raw feature extractors, \ie how they ``see'' objects, as well as a proxy for final instance segmentation performance.
Our analysis indicates that features generated by MAE lead to an overall higher recall when multiple instances exist, \ie features produced by MAE are better at discriminating multiple instances of the same semantic class, while DINO features are superior at detecting single instances. 

\begin{figure*}[p]
    \clearpage
    \small
    \centering
    \begin{tabular}{@{}l@{\hskip 4pt}c@{\hskip 4pt}cc c@{\hskip 4pt}c@{\hskip 4pt}c@{\hskip 4pt}c@{\hskip 4pt}c@{}}
        \toprule
         &\multirow{2}{*}[-0.5em]{Image} & \multirow{2}{*}[-0.5em]{GT Annotations} & \multirow{2}{*}[-0.5em]{Features} & \multicolumn{5}{c}{k} \\
         \cmidrule(l){5-9}
         {} & {} & {} & {} & 2 & 3 & 4 & 5 & 6 \\
         \midrule
         \multirow{2}{*}{(a)} & \multirow{2}{*}[0.75em]{\includegraphics[width=.11\textwidth]{images/supplementary/2006/image.jpg}} & \multirow{2}{*}[0.75em]{\includegraphics[width=.11\textwidth]{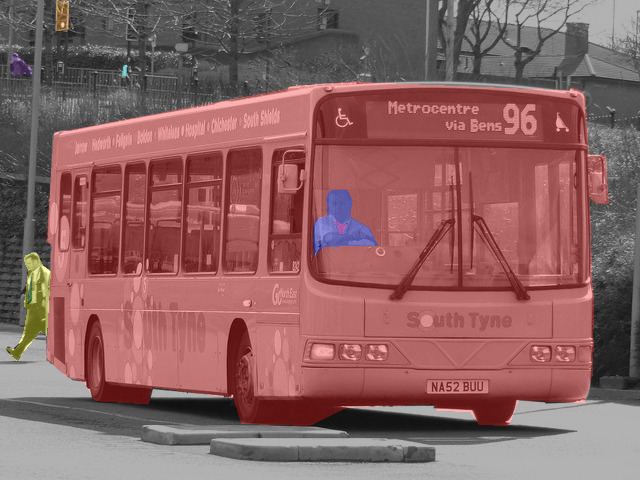}} & \raisebox{1.25em}{MAE} & \includegraphics[width=.1\textwidth]{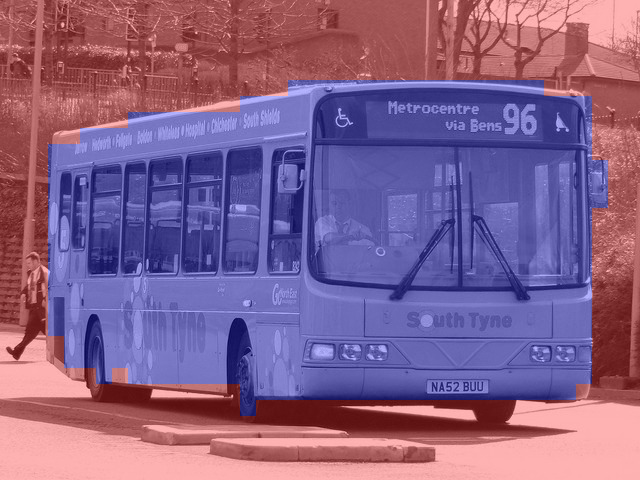} & \includegraphics[width=.1\textwidth]{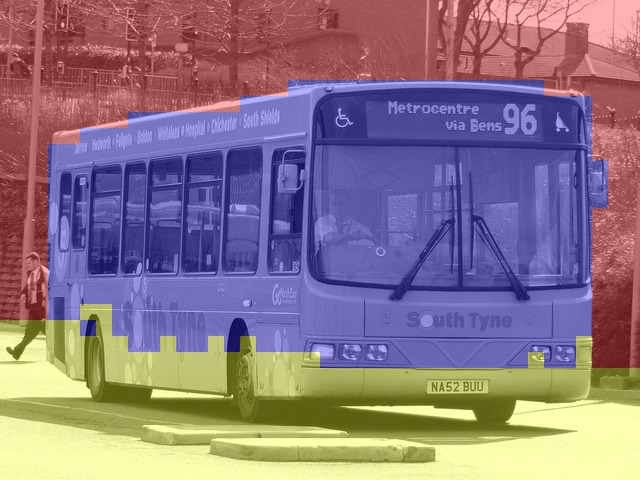} & \includegraphics[width=.1\textwidth]{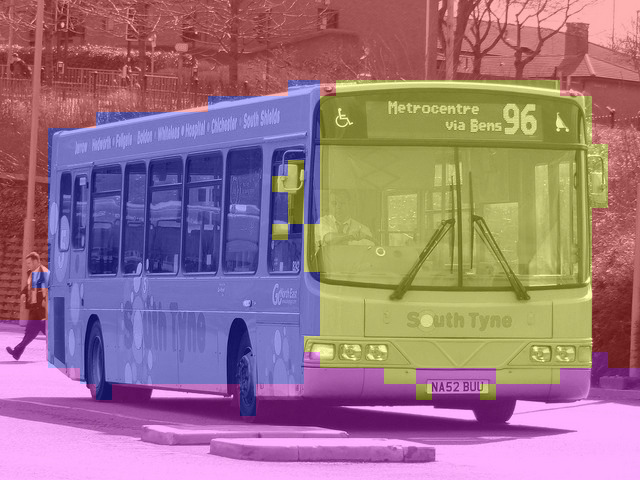} & \includegraphics[width=.1\textwidth]{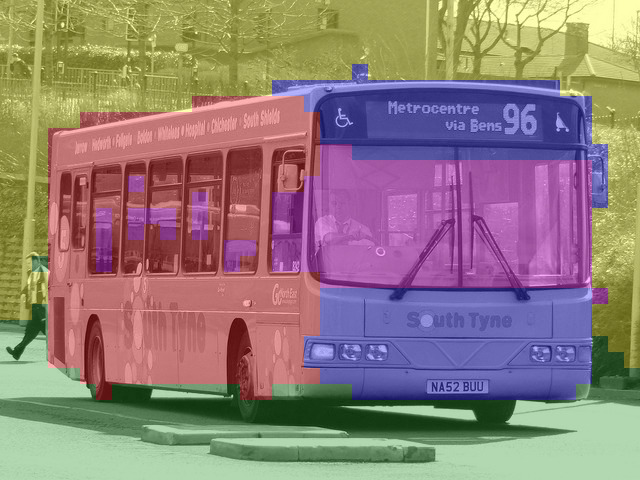}  & \includegraphics[width=.1\textwidth]{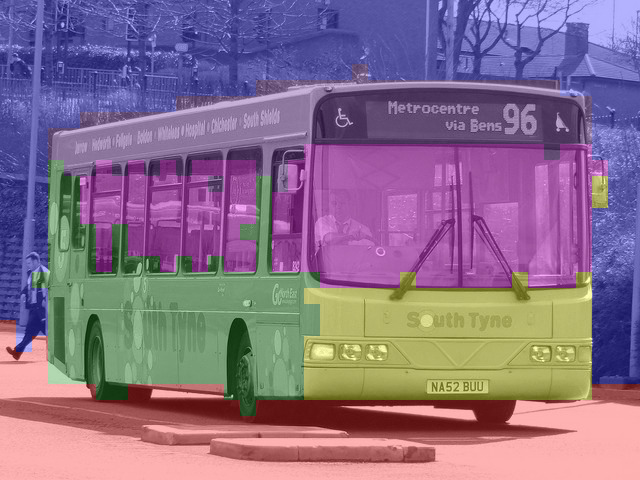} \\
         {} & {} & {} & \raisebox{1.35em}{DINO} & \includegraphics[width=.1\textwidth]{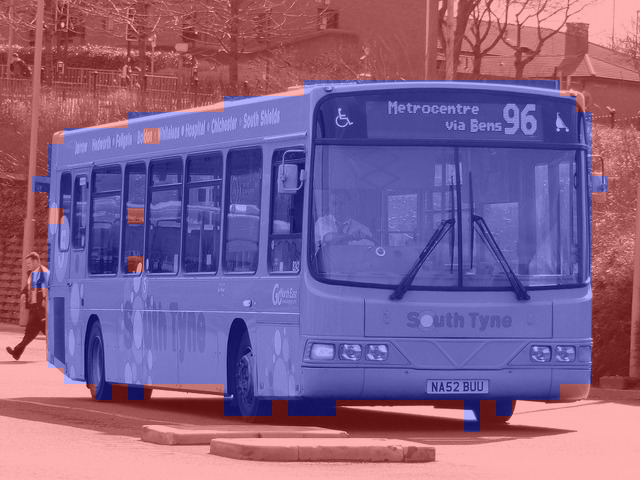} & \includegraphics[width=.1\textwidth]{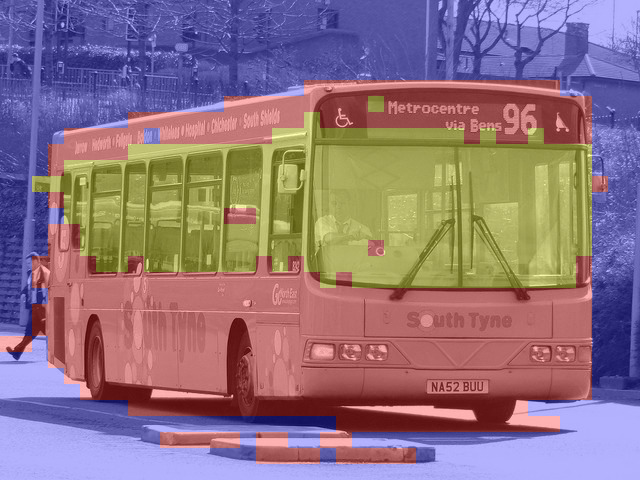} & \includegraphics[width=.1\textwidth]{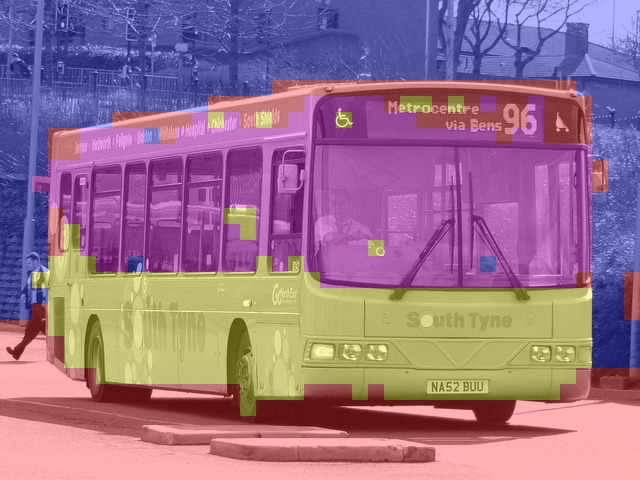} & \includegraphics[width=.1\textwidth]{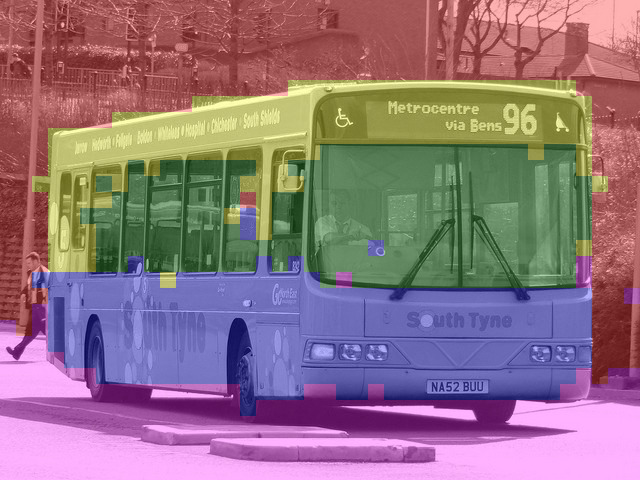}  & \includegraphics[width=.1\textwidth]{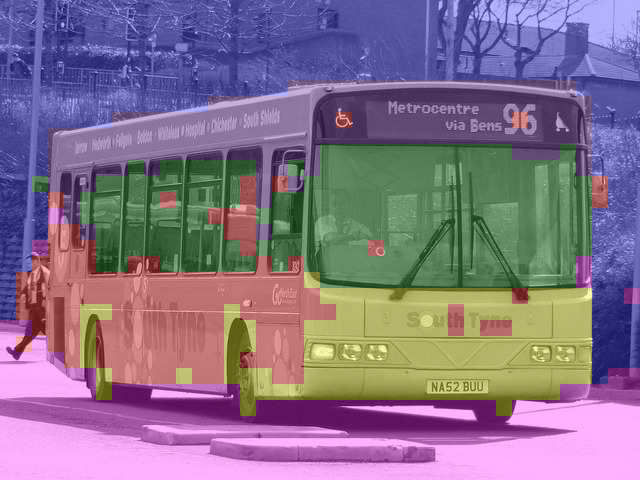} \\
         \midrule
         \multirow{2}{*}{(b)} & \multirow{2}{*}[0.75em]{\includegraphics[width=.11\textwidth]{images/supplementary/52462/image.jpg}} & \multirow{2}{*}[0.75em]{\includegraphics[width=.11\textwidth]{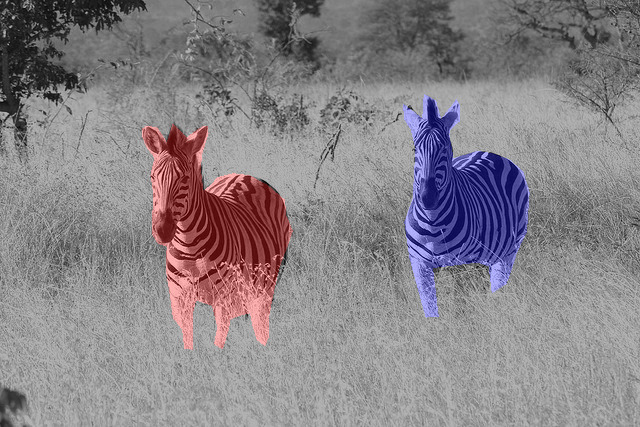}} & \raisebox{1.25em}{MAE} & \includegraphics[width=.1\textwidth]{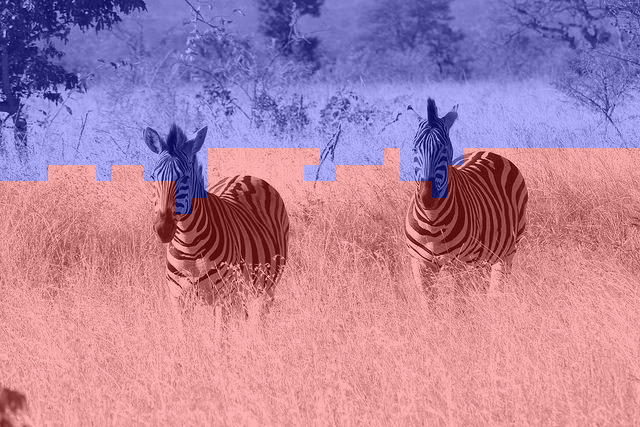} & \includegraphics[width=.1\textwidth]{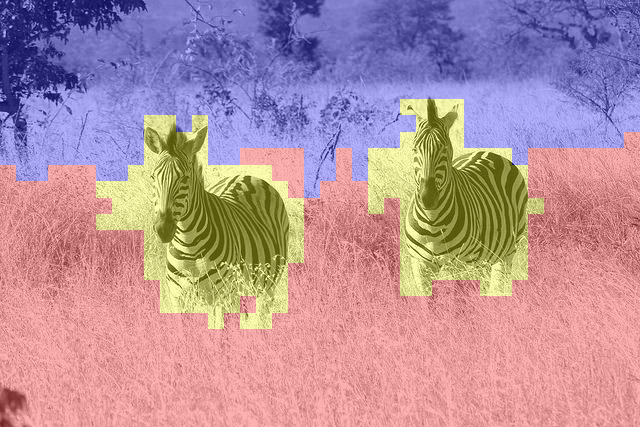} & \includegraphics[width=.1\textwidth]{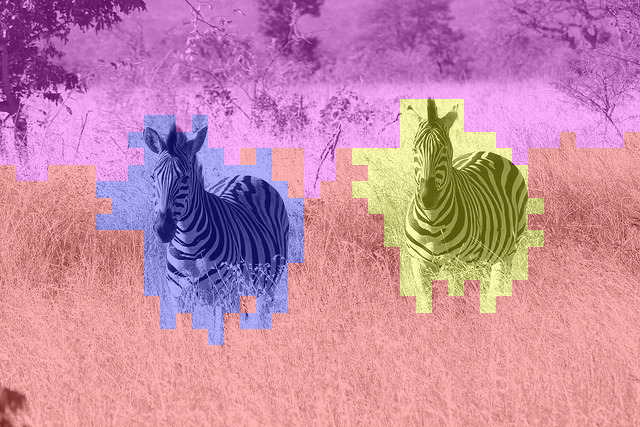} & \includegraphics[width=.1\textwidth]{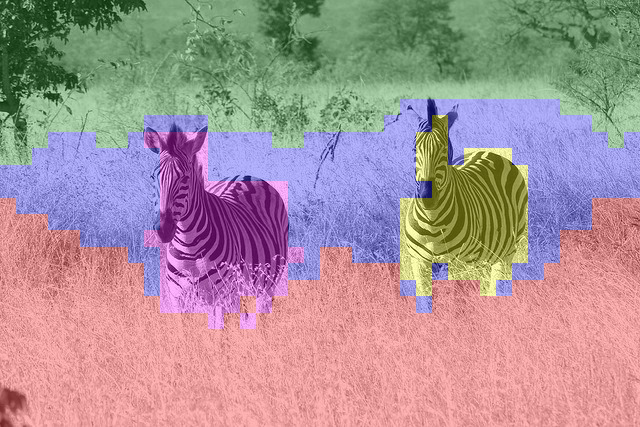}  & \includegraphics[width=.1\textwidth]{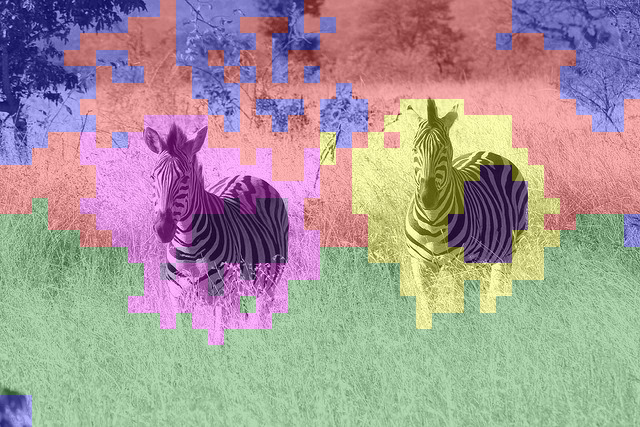} \\
         {} & {} & {} & \raisebox{1.35em}{DINO} & \includegraphics[width=.1\textwidth]{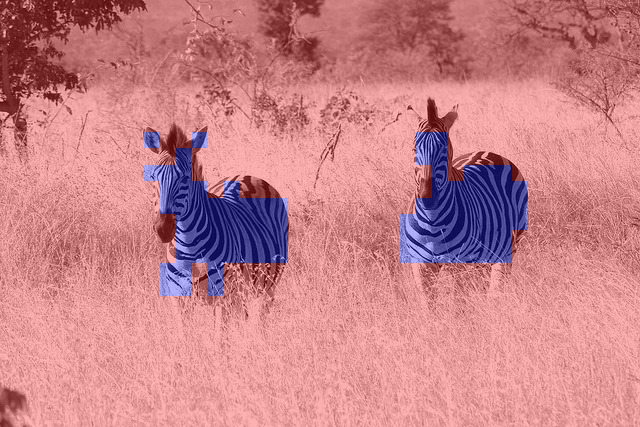} & \includegraphics[width=.1\textwidth]{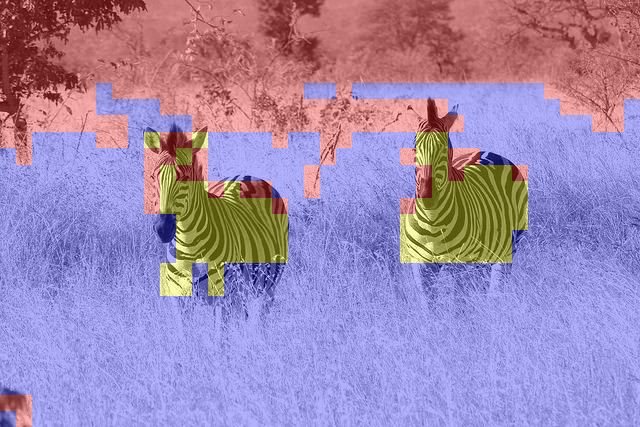} & \includegraphics[width=.1\textwidth]{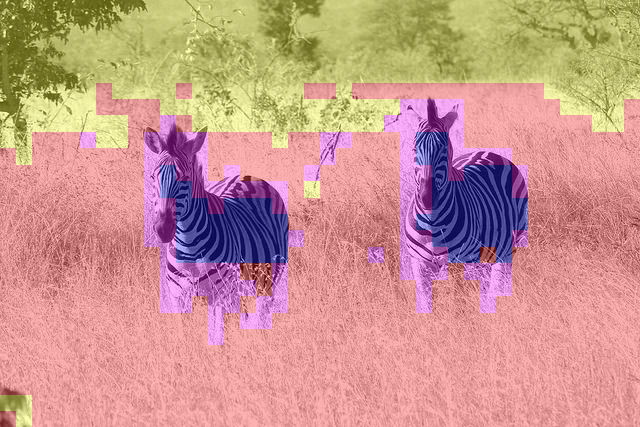} & \includegraphics[width=.1\textwidth]{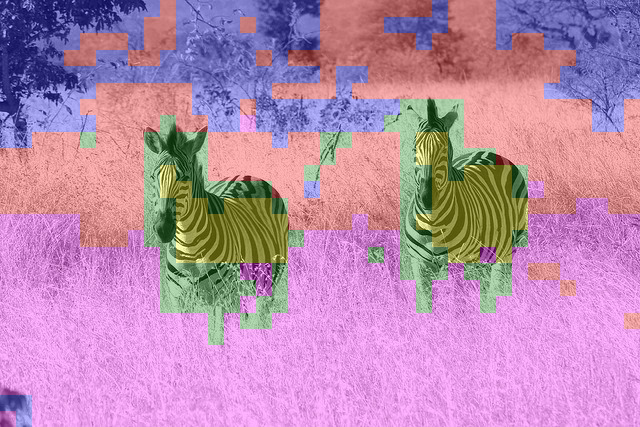}  & \includegraphics[width=.1\textwidth]{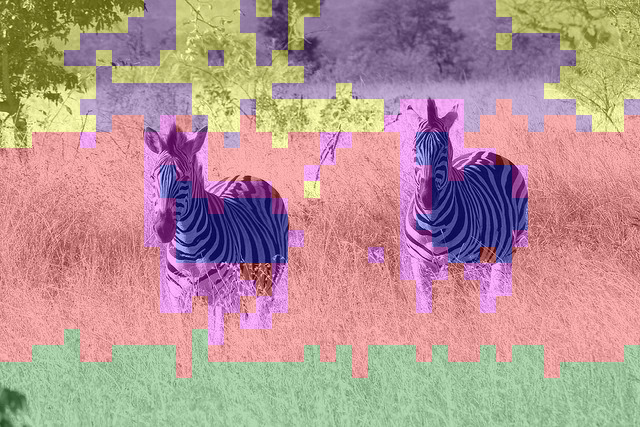} \\
         \midrule
         \multirow{2}{*}{(c)} & \multirow{2}{*}[0.75em]{\includegraphics[width=.11\textwidth]{images/supplementary/79229/image.jpg}} & \multirow{2}{*}[0.75em]{\includegraphics[width=.11\textwidth]{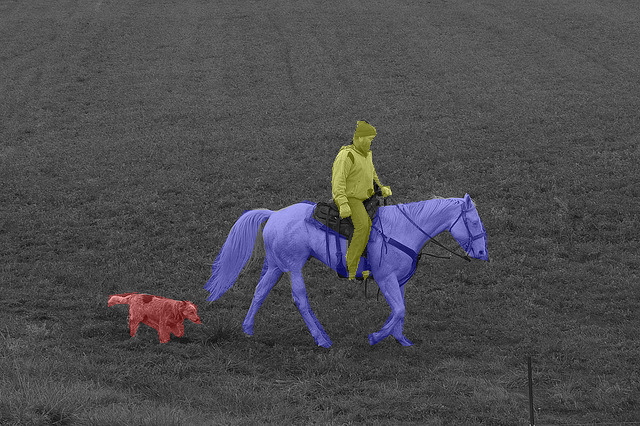}} & \raisebox{1.25em}{MAE} & \includegraphics[width=.1\textwidth]{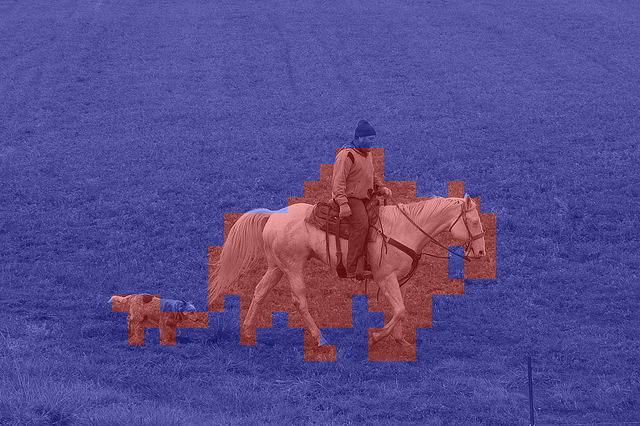} & \includegraphics[width=.1\textwidth]{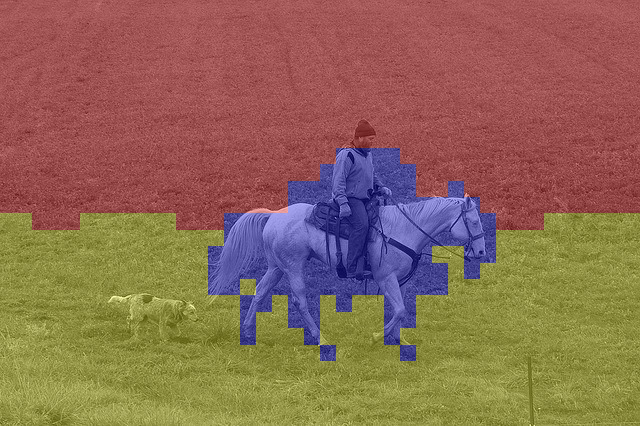} & \includegraphics[width=.1\textwidth]{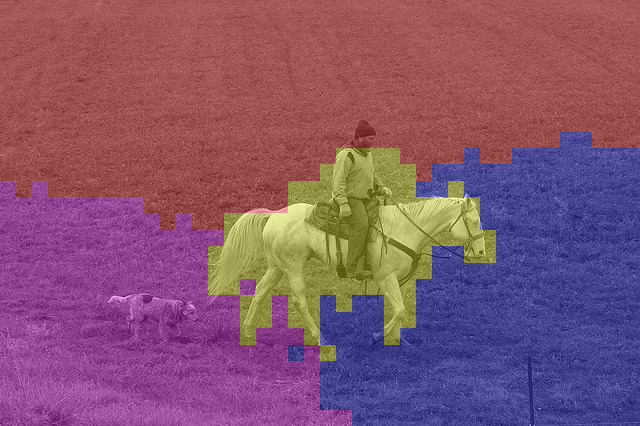} & \includegraphics[width=.1\textwidth]{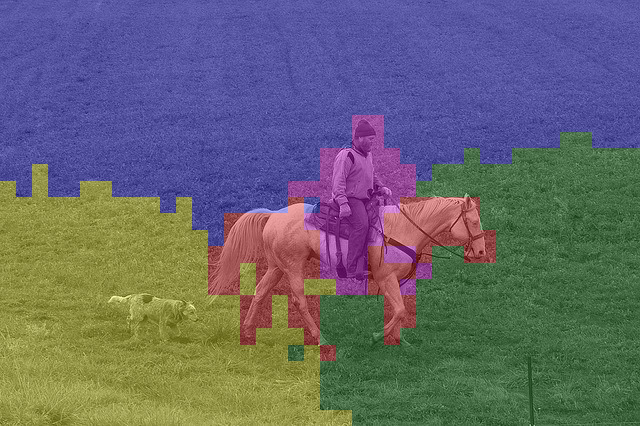}  & \includegraphics[width=.1\textwidth]{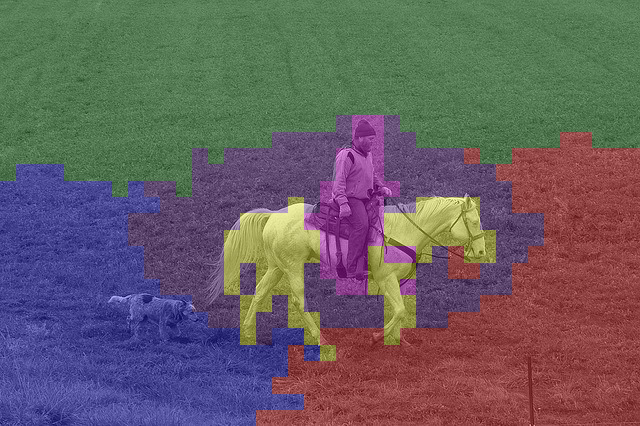} \\
         {} & {} & {} & \raisebox{1.35em}{DINO} & \includegraphics[width=.1\textwidth]{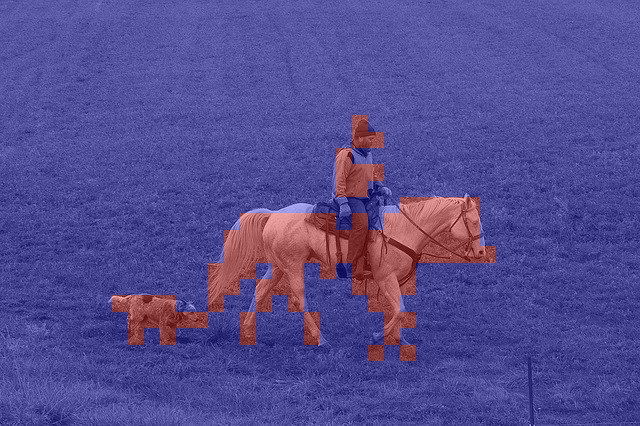} & \includegraphics[width=.1\textwidth]{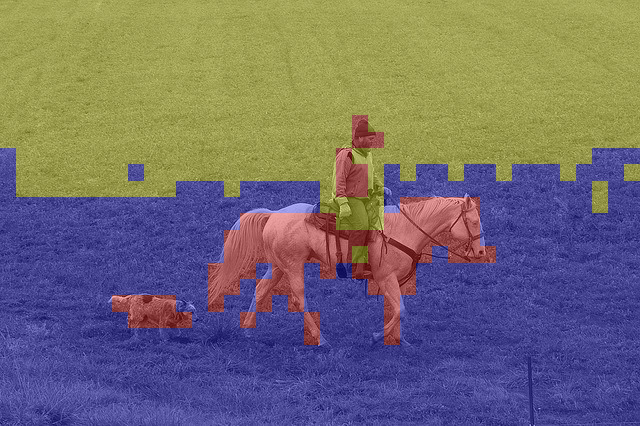} & \includegraphics[width=.1\textwidth]{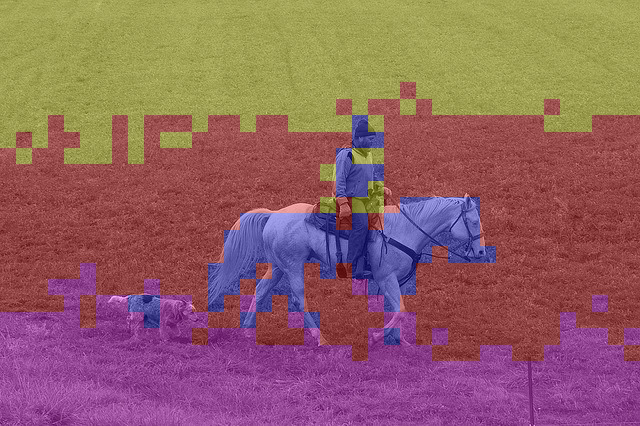} & \includegraphics[width=.1\textwidth]{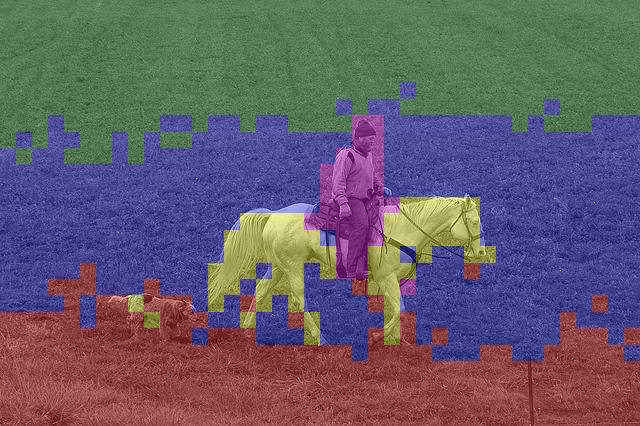}  & \includegraphics[width=.1\textwidth]{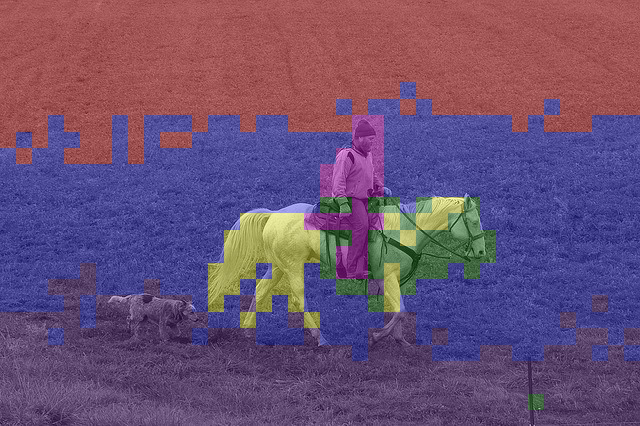} \\
         \midrule
         \multirow{2}{*}{(d)} & \multirow{2}{*}[0.75em]{\includegraphics[width=.11\textwidth]{images/supplementary/116068/image.jpg}} & \multirow{2}{*}[0.75em]{\includegraphics[width=.11\textwidth]{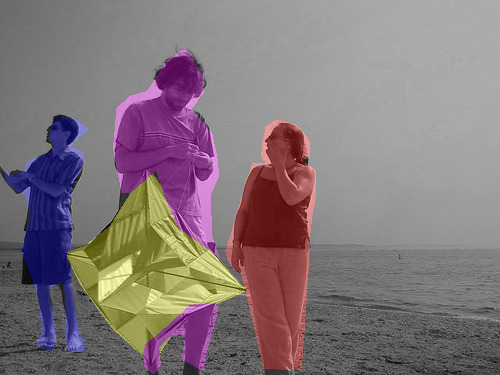}} & \raisebox{1.25em}{MAE} & \includegraphics[width=.1\textwidth]{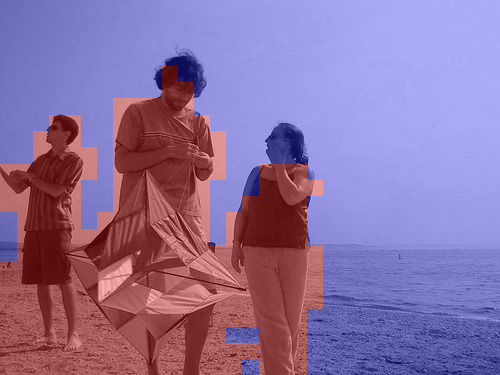} & \includegraphics[width=.1\textwidth]{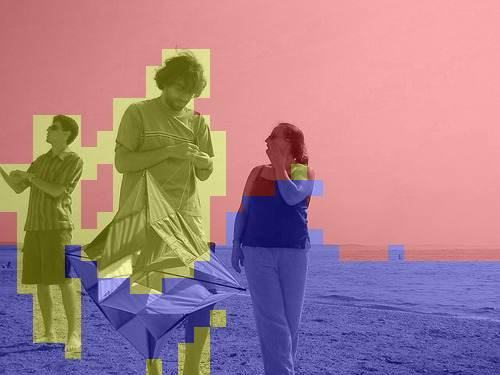} & \includegraphics[width=.1\textwidth]{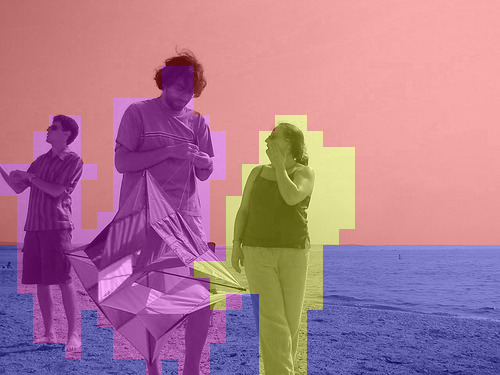} & \includegraphics[width=.1\textwidth]{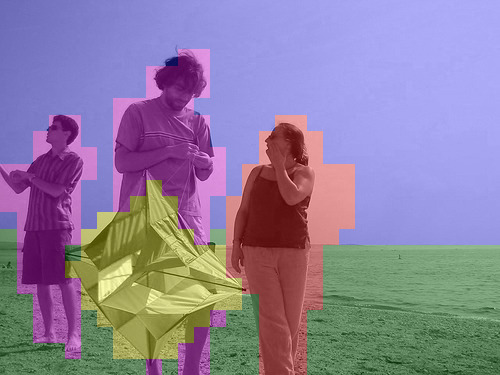}  & \includegraphics[width=.1\textwidth]{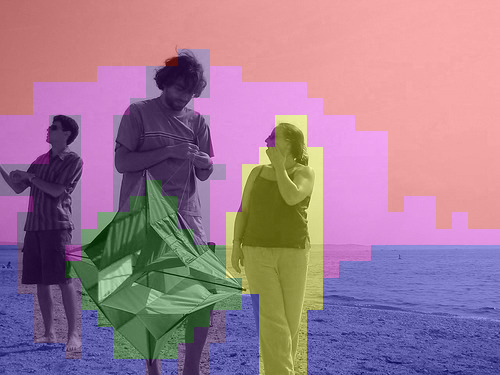} \\
         {} & {} & {} & \raisebox{1.35em}{DINO} & \includegraphics[width=.1\textwidth]{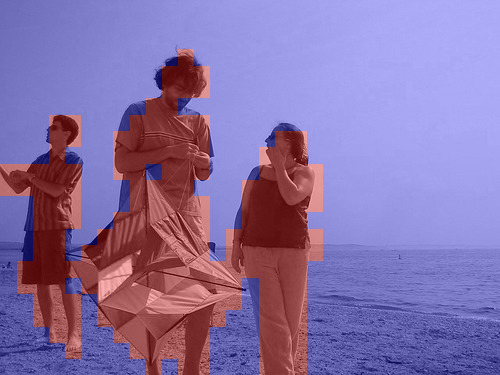} & \includegraphics[width=.1\textwidth]{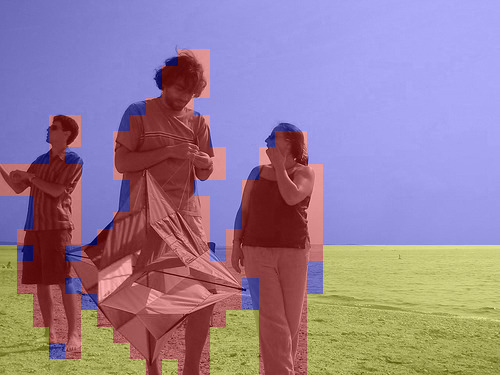} & \includegraphics[width=.1\textwidth]{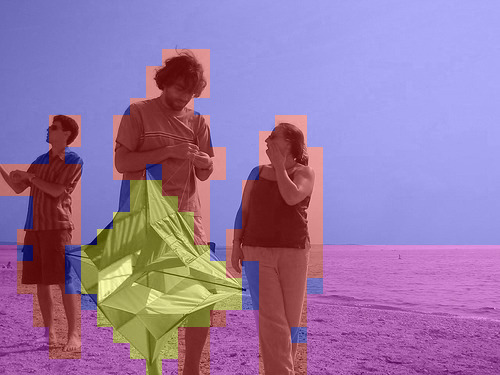} & \includegraphics[width=.1\textwidth]{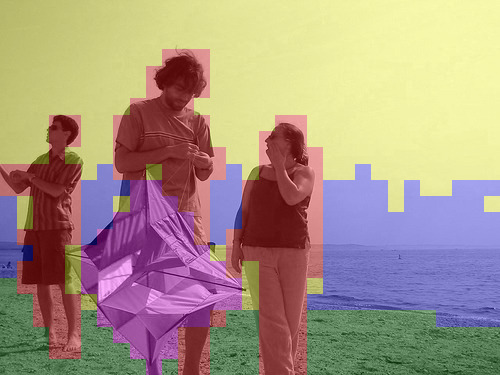}  & \includegraphics[width=.1\textwidth]{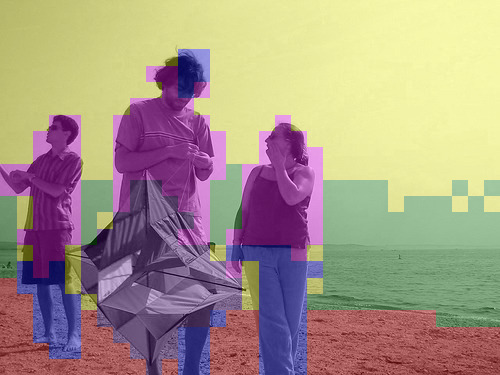} \\
         \midrule
         \multirow{2}{*}{(e)} & \multirow{2}{*}[0.75em]{\includegraphics[width=.11\textwidth]{images/supplementary/283070/image.jpg}} & \multirow{2}{*}[0.75em]{\includegraphics[width=.11\textwidth]{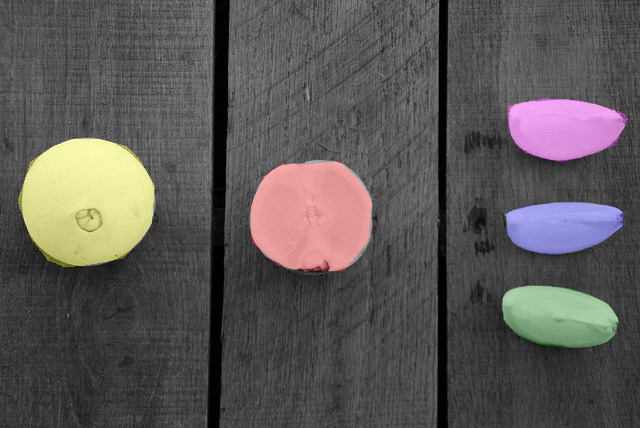}} & \raisebox{1.25em}{MAE} & \includegraphics[width=.1\textwidth]{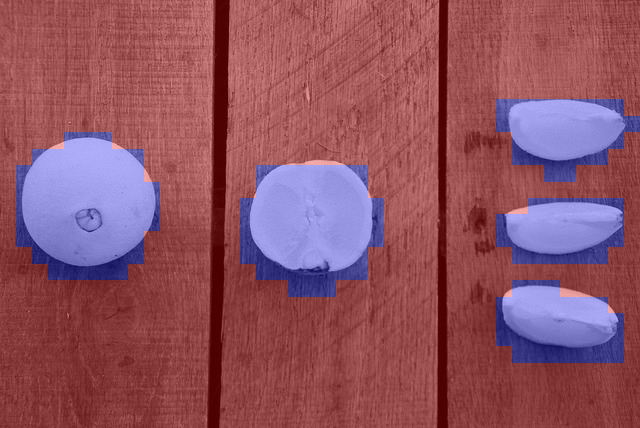} & \includegraphics[width=.1\textwidth]{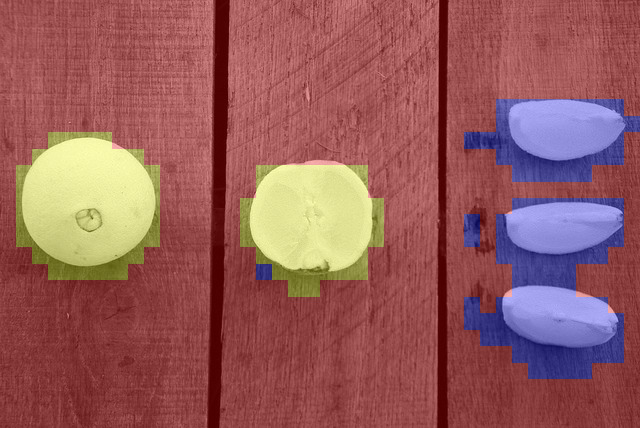} & \includegraphics[width=.1\textwidth]{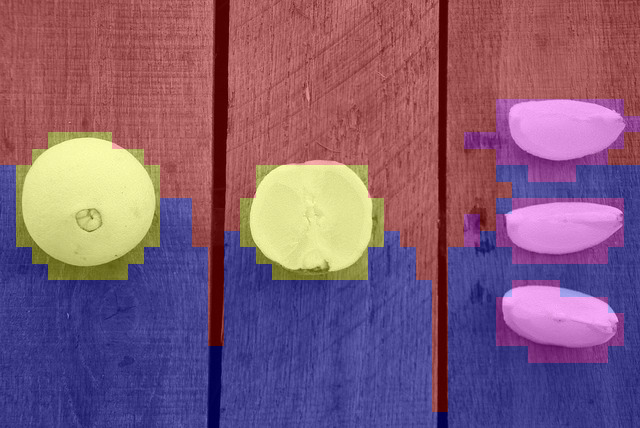} & \includegraphics[width=.1\textwidth]{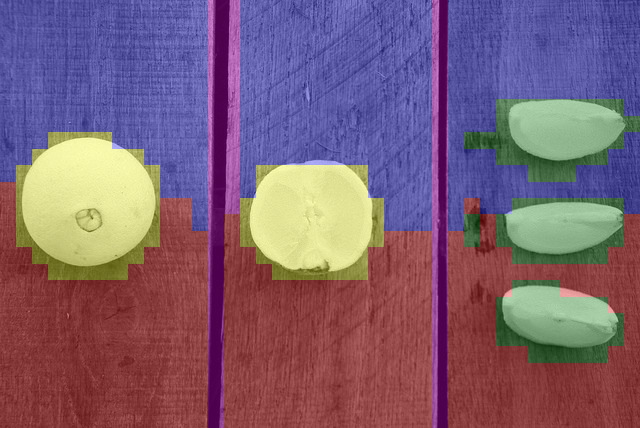}  & \includegraphics[width=.1\textwidth]{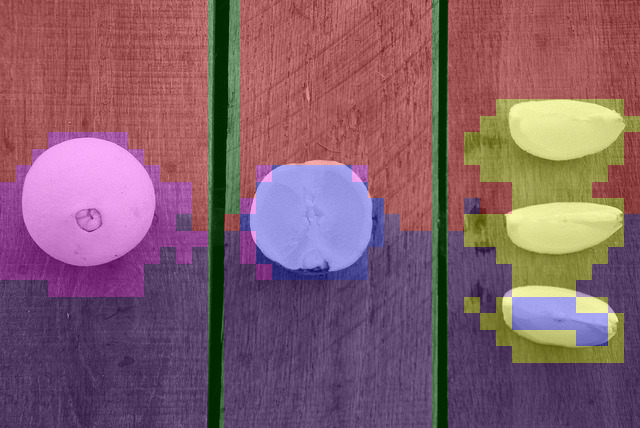} \\
         {} & {} & {} & \raisebox{1.35em}{DINO} & \includegraphics[width=.1\textwidth]{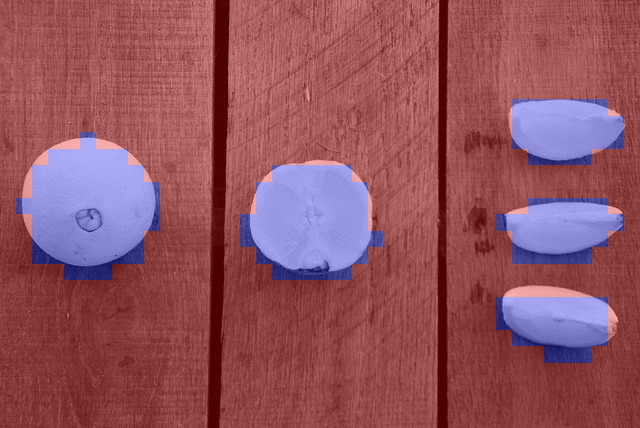} & \includegraphics[width=.1\textwidth]{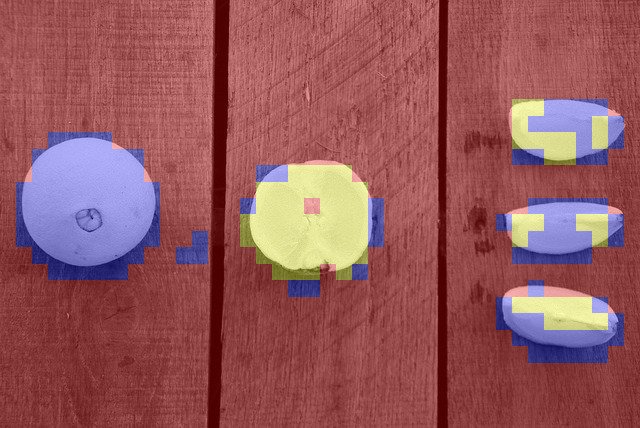} & \includegraphics[width=.1\textwidth]{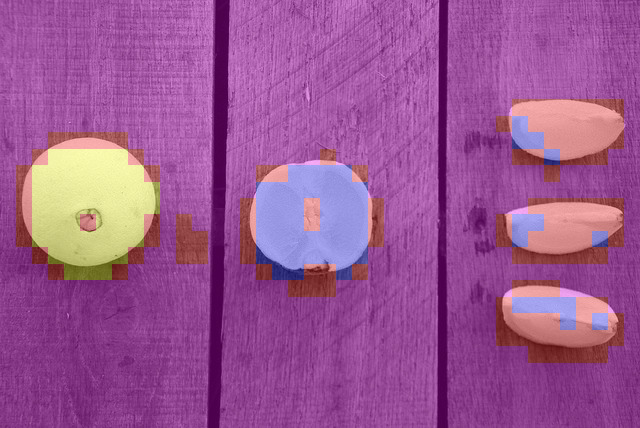} & \includegraphics[width=.1\textwidth]{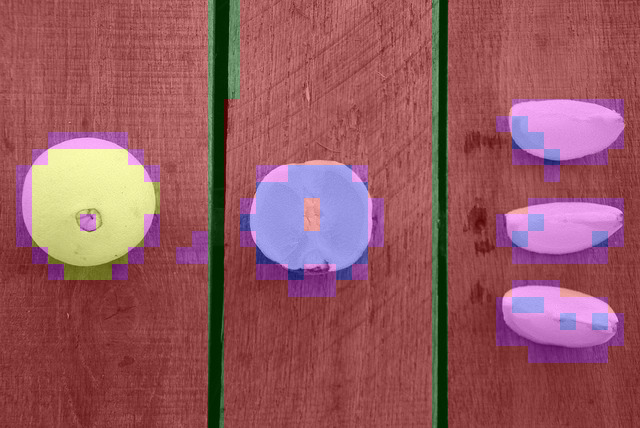}  & \includegraphics[width=.1\textwidth]{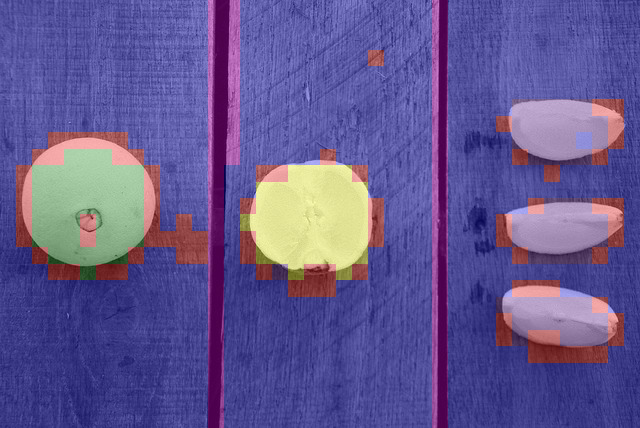} \\
         \midrule
         \multirow{2}{*}{(f)} & \multirow{2}{*}[0.75em]{\includegraphics[width=.11\textwidth]{images/supplementary/123213/image.jpg}} & \multirow{2}{*}[0.75em]{\includegraphics[width=.11\textwidth]{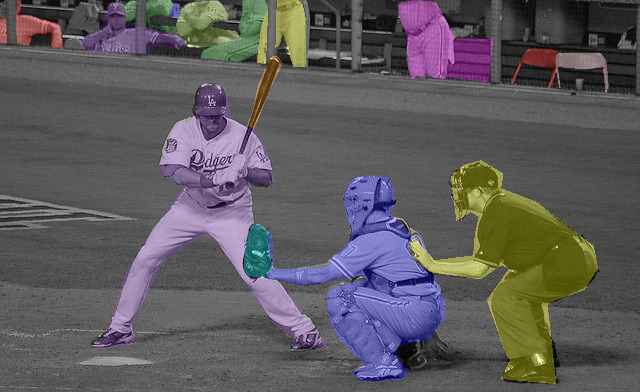}} & \raisebox{1.25em}{MAE} & \includegraphics[width=.1\textwidth]{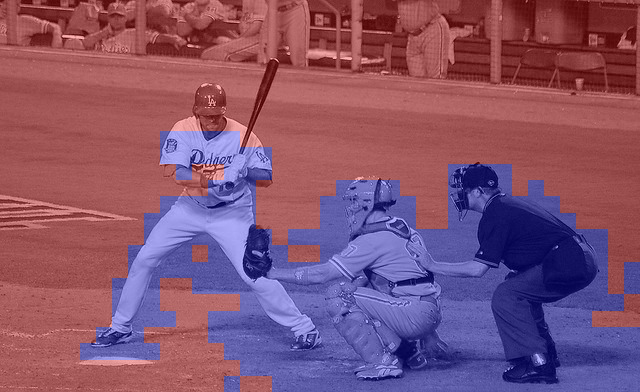} & \includegraphics[width=.1\textwidth]{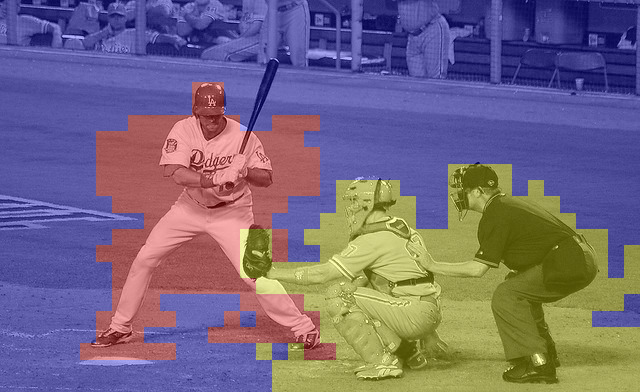} & \includegraphics[width=.1\textwidth]{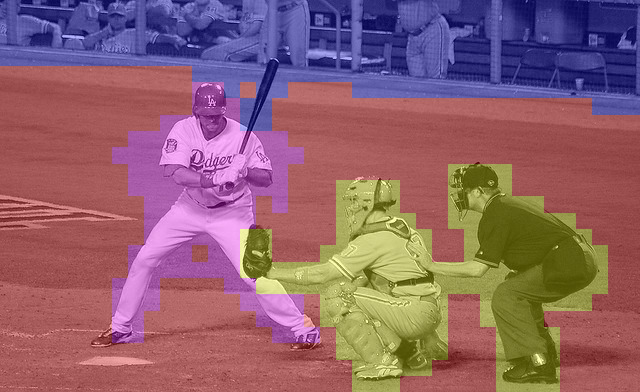} & \includegraphics[width=.1\textwidth]{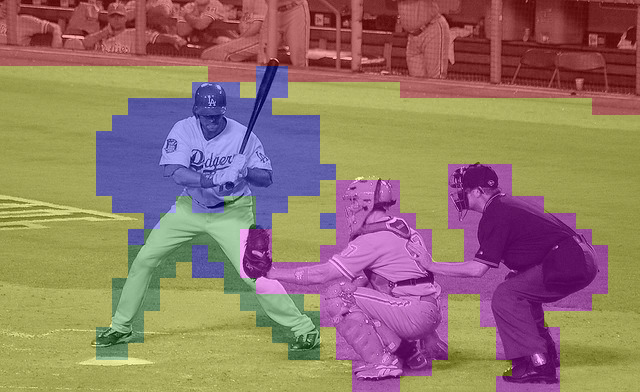}  & \includegraphics[width=.1\textwidth]{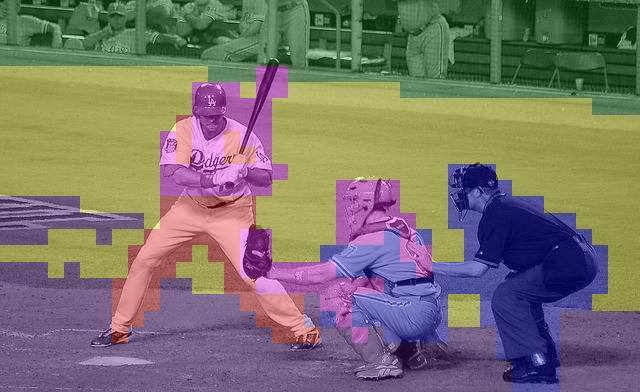} \\
         {} & {} & {} & \raisebox{1.35em}{DINO} & \includegraphics[width=.1\textwidth]{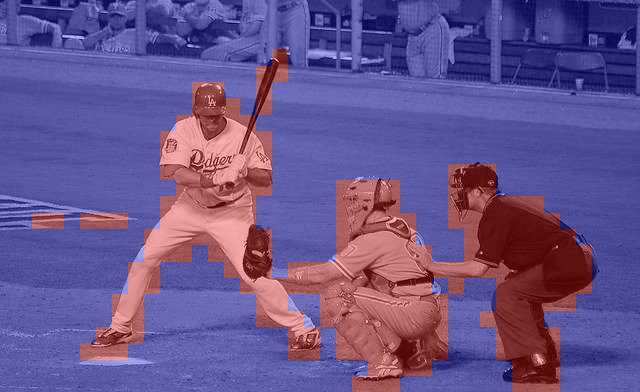} & \includegraphics[width=.1\textwidth]{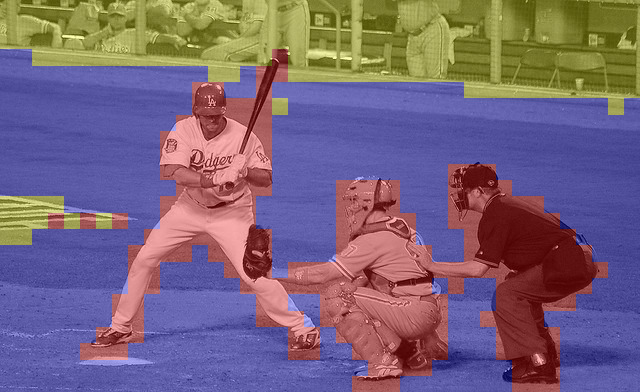} & \includegraphics[width=.1\textwidth]{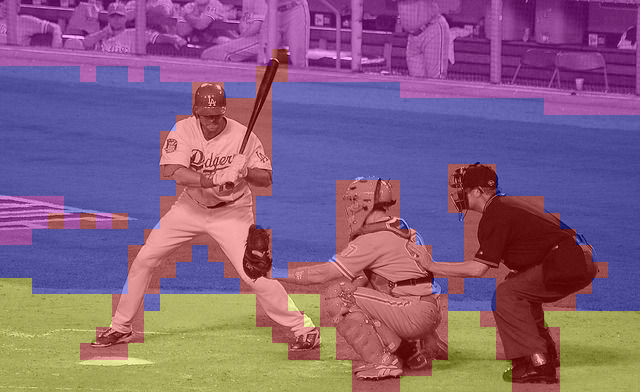} & \includegraphics[width=.1\textwidth]{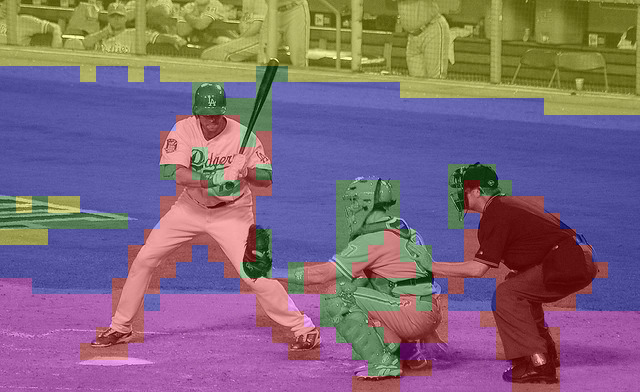}  & \includegraphics[width=.1\textwidth]{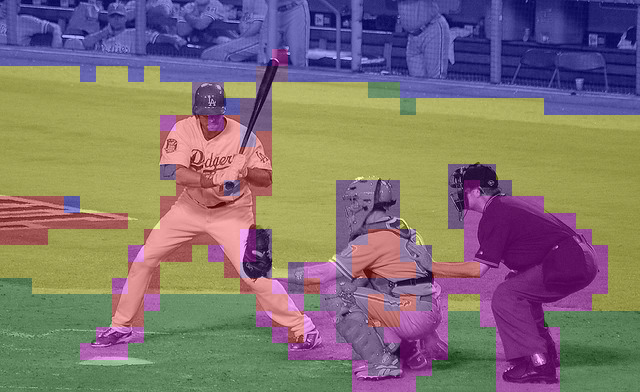} \\
         \midrule
         \multirow{2}{*}{(g)} & \multirow{2}{*}[0.75em]{\includegraphics[width=.11\textwidth]{images/supplementary/563653/image.jpg}} & \multirow{2}{*}[0.75em]{\includegraphics[width=.11\textwidth]{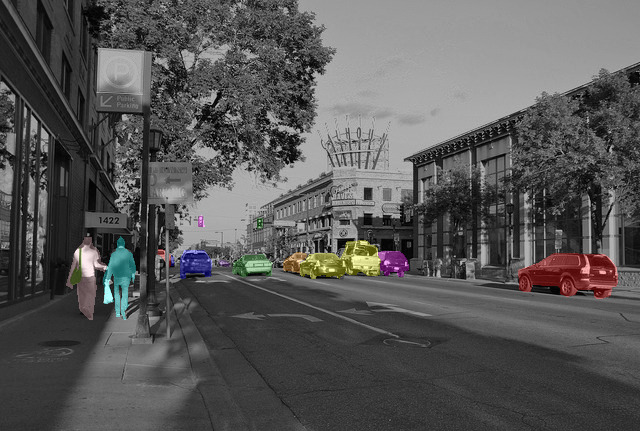}} & \raisebox{1.25em}{MAE} & \includegraphics[width=.1\textwidth]{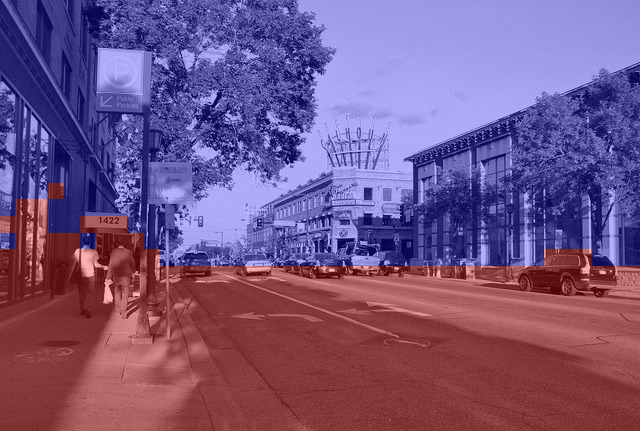} & \includegraphics[width=.1\textwidth]{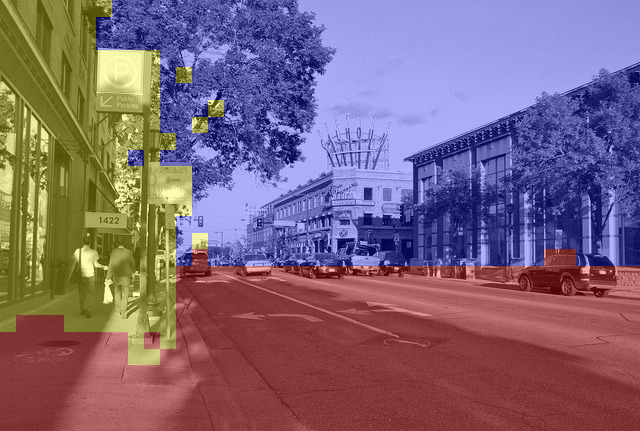} & \includegraphics[width=.1\textwidth]{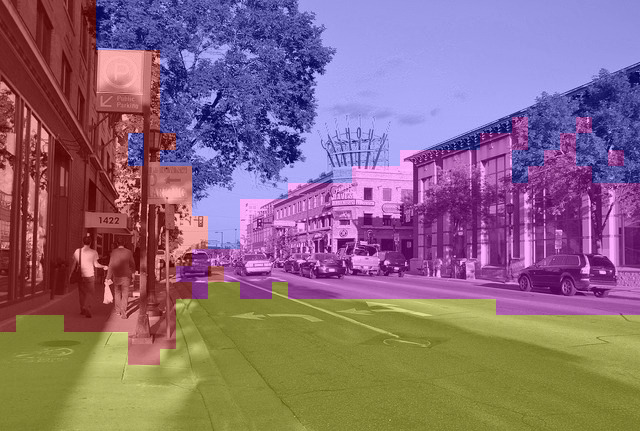} & \includegraphics[width=.1\textwidth]{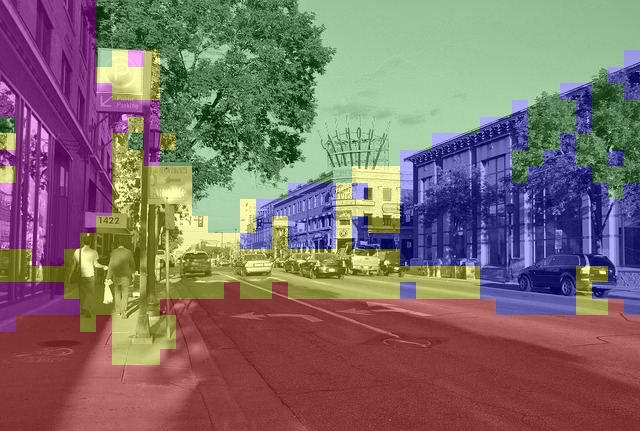}  & \includegraphics[width=.1\textwidth]{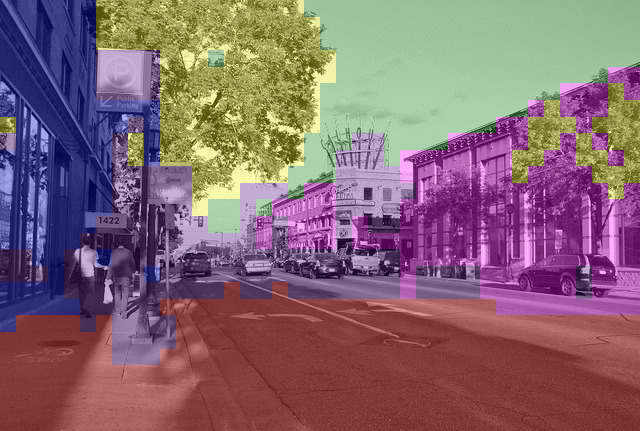} \\
         {} & {} & {} & \raisebox{1.35em}{DINO} & \includegraphics[width=.1\textwidth]{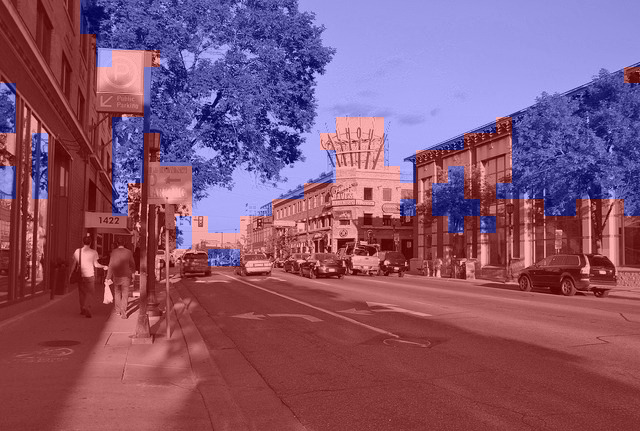} & \includegraphics[width=.1\textwidth]{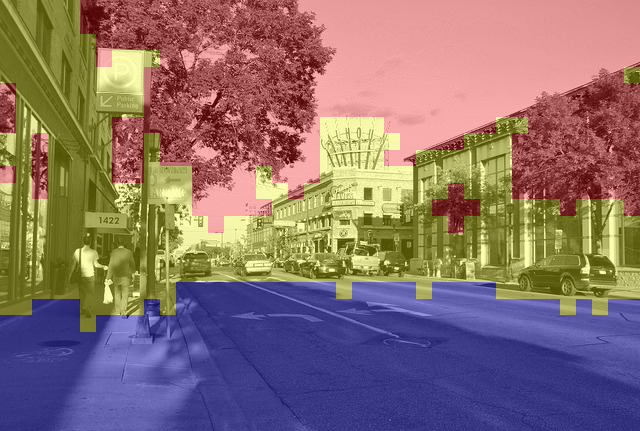} & \includegraphics[width=.1\textwidth]{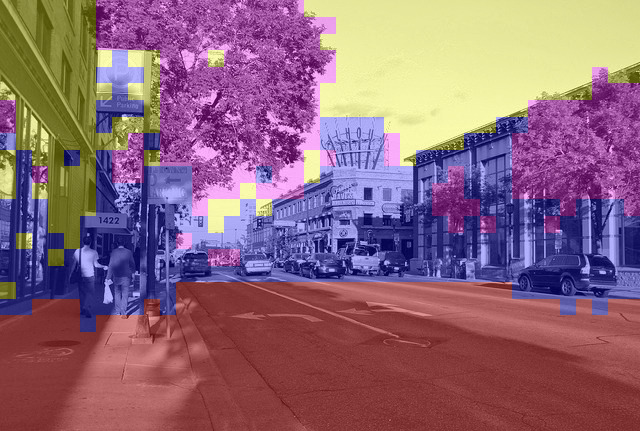} & \includegraphics[width=.1\textwidth]{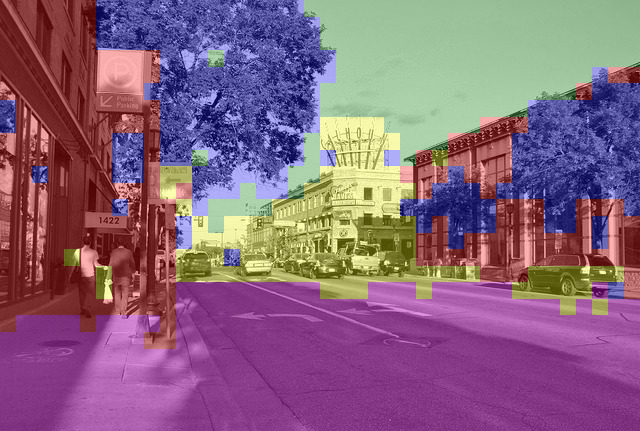}  & \includegraphics[width=.1\textwidth]{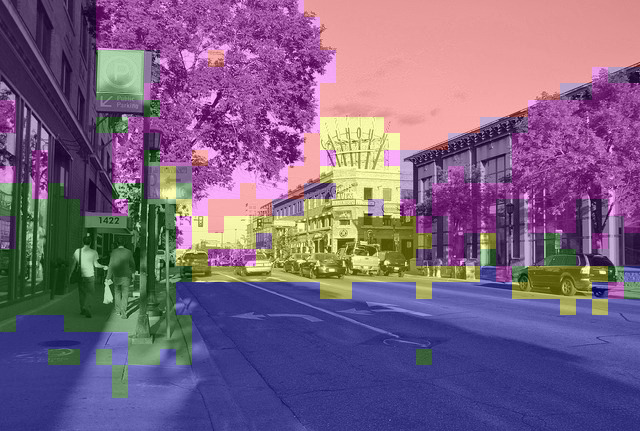} \\
        \bottomrule
    \end{tabular}
    \caption{\textbf{Spectral clustering of self-supervised features from MAE and DINO with $k \in \{2, ..., 6\}$.} We show a sample of images from COCO \texttt{val2017} with different numbers of ground-truth (GT) annotations. For each feature extractor, we show all masks for individual values of $k$. Each mask is shown with a different color, where \textit{no color} represents the absence of annotations. Best viewed on screen.}
    \label{tab:mini-experiment-images}
\clearpage
\end{figure*}

We illustrate this observation further in \cref{tab:mini-experiment-images} with qualitative examples.  
We show examples from COCO \texttt{val2017} and compare the segments produced by MAE and DINO after spectral clustering while varying $k$.

Next, we record some qualitative observations from our experience with these methods, which we hope will be of use to researchers working with them in the future:

\begin{itemize}
\item Spectral clustering of MAE features has a tendency to separate instances of the same semantic class in the same image, whereas DINO features are more likely to produce a purely semantic grouping (\eg, \cref{tab:mini-experiment-images} (b)). 
Yet there is ambiguity as to how semantic classes are ``seen'' by self-supervised models. 
An example is shown in \cref{tab:mini-experiment-images} (e), in which all objects correspond to the same semantic class \textit{orange}. 
The image shows a whole orange, a halved orange, and three individual peeled segments, which represent different \textit{states} of the orange.
Since these states have distinct appearances, DINO is able to partially discriminate between these objects, and its behavior comes closer to MAE in resembling ``instance awareness'' in this example.
This finding aligns with the observation of \cite{laina2022measuring} and is likely true for most self-supervised models.   

\item{For smaller values of $k$ (especially if $k$ is smaller than the number of objects in the image), the decomposition remains shallow. Even if $k$ is sufficiently large to capture all objects, a number of clusters are often used to represent the nuances of the background, and as a result, different foreground objects group together in the remaining clusters.} 

\item With an increasing number of clusters (over-clustering), MAE features tend to \emph{spatially} decompose an image, whereas, with DINO, the level of semantic detail seems to increase, \eg objects get subdivided into parts (\cref{tab:mini-experiment-images} (a), (b), (d)). % (\eg, trees are lifted from the background).

\item Applying spectral clustering to DINO features with $k=2$ produces higher quality saliency candidates (\cref{tab:mini-experiment-images} (b), (f)), which matches the results of the ablation in Tab.~7.

\item Both feature extractors struggle with very small objects, \eg the dog (\cref{tab:mini-experiment-images} (c)) or the baseball bat and glove (\cref{tab:mini-experiment-images} (f)). In these examples, small objects become part of larger, more prominent objects or part of the background.

\item Complex scenes without prominent foreground objects, \eg scenes ``seen'' from a distance (\cref{tab:mini-experiment-images} (g)), yield % exclusively bad masks (at least in terms of valid COCO classes)
segmentations that do not align with the valid object categories in COCO\,---\,for example, the image is partitioned into trees, buildings, and street.
\end{itemize}

\subsection{Self-Training for Instance Segmentation, Addendum}
\subsubsection{Datasets and Training Details}
\label{ss:app:details}

\begin{table*}
  \footnotesize
  \centering
  \def\arraystretch{0.95}
  \setlength{\tabcolsep}{5pt}
   \caption{\textbf{Unsupervised class-agnostic instance segmentation.} 
  We report the performance of state-of-the-art methods on this task on common benchmarks. For FreeSOLO, we report additional results marked with $\dagger$ based on our reproduction. FreeSOLO and Exemplar-FreeSOLO use DenseCL features, while MaskDistill and CutLER use DINO. Our setting is most comparable to FreeSOLO.
  }
  \begin{tabular}{lcccccccccccc}
  \toprule
  & \multicolumn{3}{c}{\textbf{PASCAL VOC}} & \multicolumn{3}{c}{\textbf{COCO20k}} & \multicolumn{3}{c}{\textbf{COCO (val2017)}} & \multicolumn{3}{c}{\textbf{UVO}} \\
  \cmidrule(lr){2-4} \cmidrule(lr){5-7} \cmidrule(lr){8-10} \cmidrule(lr){11-13} 
  \textbf{Method} & AP$_\text{50}$ & AP$_\text{75}$ & AP & AP$_\text{50}$ & AP$_\text{75}$ & AP & AP$_\text{50}$ & AP$_\text{75}$ & AP & AP$_\text{50}$ & AP$_\text{75}$ & AP \\
  \midrule
  FreeSOLO~\cite{wang2022freesolo} & 9.8$^\dagger$ & 0.2$^\dagger$ & 2.3$^\dagger$ & 10.2$^\dagger$ & 3.5$^\dagger$ & 4.6$^\dagger$ & 9.8 & 2.9 & 4.0 & 12.7 & 3.0 & 4.8  \\
  Exemplar-FS~\cite{Ishtiak_2023_CVPR} & - & - & - & - & - & - & 13.2 & 6.3 & 8.4 & 14.2 & 7.3 & 9.2 \\
  MaskDistill~\cite{van2022discovering} & 24.3 & 6.9 & 9.9 & 6.8 & 2.1 & 2.9 & - & - & - & - & - & - \\
  CutLER~\cite{wang2023cut} & - & - & - & 19.6 & 10.0 & 9.2 & 18.9 & 9.7 & 9.2 & 22.8 & 8.0 & 10.1 \\
  \midrule
  Ours (MAE) & 18.8 & 0.6 & 4.6 & 12.3 & 3.7 & 5.3 & 12.0 & 3.7 & 5.2 & 12.9 & 2.6 & 4.7  \\
  \bottomrule
  \end{tabular}
  \label{table:seg-all}
\end{table*}

\paragraph*{MS COCO 2017.} The \textbf{C}ommon \textbf{O}bjects in \textbf{CO}ntext dataset comprises images of ``complex everyday scenes containing common objects in their natural context"~\cite{lin2014microsoft}. 
We use the 2017 release in our method, which consists of \texttt{train2017}, \texttt{unlabeled2017}, and \texttt{val2017} sets. 
The training set contains 118,287 images with 860,001 polygon annotations for 80 semantic classes sourced from human annotators. 
The unlabeled image set comprises 123,403 images without any annotations.
We use the \texttt{train2017} and \texttt{unlabeled2017} sets for generating mask proposals with our method to bootstrap an instance segmentation network (no ground truth annotations are used). 
We then evaluate it on \texttt{val2017}, which contains 5,000 images with 36,781 polygon annotations, in a class-agnostic (\ie, classes ignored) setting.

\paragraph*{Training Details.}
We generate the mask proposals using pre-trained ViT-B/16 vision transformers (except for SelfPatch, which is only available for ViT-S/16, and DenseCL, which is a ResNet-101 model) and a ViT-B/16 DINO model for saliency estimation. 
These networks are all pre-trained on ImageNet \cite{ILSVRC15} without supervision.
When extracting features, we do not apply any data augmentation to the images.
We choose $K = \{2, 3, 4, 5\}$ as the set of values for $k$ to generate candidate masks and further split non-connected components into separate masks.
To obtain the final set of mask proposals, we first discard masks with an intersection score of less than 0.5 when compared to the saliency map. 
Then, we apply NMS with a threshold of $0.8$ to remove any duplicates from the masks that remain.

Unless otherwise specified, we use SOLOv2~\cite{wang2022freesolo,wang2020solov2} as the instance segmentation network, initialized with self-supervised weights (DenseCL, ResNet-101). We employ copy \& paste augmentation as well as the BoxInst loss \cite{tian2021boxinst,wang2022freesolo}.
We train the network with a frozen backbone for 60k steps with our mask proposals as targets. 
The stochastic gradient descent optimizer is used with a learning rate of 0.00025 and a batch size of 4. 
We follow the default parameters and loss setup of \cite{wang2022freesolo}, and perform only a single round of training (which differs from \cite{wang2022freesolo} that employs two rounds with 30k steps each).
The trained network returns a confidence score for each detection/mask, which allows us to set a threshold and retain the most confident predictions. 
Training takes 48 hours on an NVIDIA A40 GPU.

\subsubsection{State-Of-The-Art on Unsupervised Instance Segmentation}

In \cref{table:seg-all}, we present the performance of recent works for the task of unsupervised class-agnostic instance segmentation and provide a comparison to our simple approach (using MAE features to generate mask proposals).

Apart from COCO \texttt{val2017} (discussed above), we show results for:
\begin{enumerate}
    \item PASCAL VOC, a standard benchmark for objection detection with annotations spanning 20 object categories that overlap with the COCO categories. It contains 5,011 images and 15,662 object annotations.
    \item COCO20k, a subset of the COCO 2014 \texttt{trainval} split, with 19,817 images (143,951 object annotations) and excludes objects flagged as \emph{crowd}.
    \item The \textbf{U}nidentified \textbf{V}ideo \textbf{O}bjects dataset, which provides a benchmark for ``open-world class-agnostic object segmentation in videos" \cite{wang2021unidentified}. It includes dense segmentation annotations of video frames from human annotators without any class information to evaluate the degree to which models recognize object categories not seen during training. 
    For evaluation, the validation set of UVO-Sparse 1.0 is used, which comprises a total of 65,588 temporally sparse frame annotations for 8,124 frames of 2,708 videos.
\end{enumerate}

As our approach follows the setup of FreeSOLO~\cite{wang2022freesolo} (architecture, loss formulation, and hyper-parameters), it permits direct comparison and shows clear improvements across datasets. Exemplar-FreeSOLO~\cite{Ishtiak_2023_CVPR}, with its addition of a randomly drawn pool of exemplars used in a contrastive learning loss, shows stronger improvements.

MaskDistill~\cite{van2022discovering} trains separate models for each dataset, as opposed to all previously mentioned methods that are solely trained on COCO \texttt{train2017} and \texttt{unlabeled2017}. Therefore, comparability is limited.

CutLER~\cite{wang2023cut} trains on ImageNet, exploiting its object-centric prior, as most images contain a single object in the center of the frame. Due to its strong instance discrimination abilities, CutLER is the current state-of-the-art method for this task.

\subsection{Ablations}
We perform various ablations to investigate the effectiveness of our design choices surrounding the self-training setup.
For simplicity, we focus on annotations produced by the best-performing feature extractor, \ie MAE features (filtered with saliency maps obtained from DINO features), instead of considering all feature extractors individually. All models are evaluated after training an instance segmentation network for 30k steps on COCO \texttt{val2017}.

\paragraph{Choice of k.}
The set $K$ of values for $k$ determines the number of objects that can be discovered but also the number of spurious masks that might appear.
For images with a single object, large values of $k$ might over-segment objects into parts, whereas for images with multiple objects, too small values might lead to masks that do not capture individual instances or fail to segment any objects at all.
In~\cref{tab:k-ablation}, we show that for complex datasets such as COCO, where images contain various numbers of objects, no single choice of $k$ emerges as the best option.
Generating masks from multiple values of $k$, however, provides an opportunity to deal with this variance as evidenced by the significantly higher recall.

\paragraph{Generating the Saliency Map.}
The saliency map, which we use to retain masks that are likely to correspond to objects in an image while discarding those that are not, is a central piece in our self-training setup. In \cref{tab:saliency}, we compare saliency maps generated from MAE and DINO features, showing that DINO features are far superior.

\paragraph{Non-Connected Component Splitting.} 
Although feature extractors yield mask proposals that segment individual instances, they may still generate mask candidates with multiple objects in a single mask, especially for smaller values of $k$ (\cref{fig:choice-of-k}). If such masks contain disconnected components, we assume they are separate objects and thus split them apart. This is particularly helpful as we apply NMS as part of our mask generation process: If a larger $k$ does indeed capture these objects separately, the split components can be deduplicated, effectively eliminating the poor original mask candidate. In \cref{tab:ncc-splitting}, we show that this splitting of non-connected components in masks is beneficial.

\paragraph{Segmentation Architecture.}
As the final part of our self-training design, we use an instance segmentation network, which refines masks and helps to increase the number of objects that are detected. We observe that the setup of SOLOv2~\cite{wang2020solov2}, adapted to unsupervised segmentation~\cite{wang2022freesolo}, is superior to an off-the-shelf Mask R-CNN for unsupervised instance segmentation, as seen in~\cref{tab:arch}.

\begin{table}[h]
    \small
    \centering
    \setlength{\tabcolsep}{5.2pt}
     \caption{\textbf{Ablation of different values for $k$.} 
    %The models are evaluated after 30k training steps on COCO \texttt{val2017}. 
    Multi-way clustering ($K=\{2,3,4,5\}$)  performs best. %We split non-connected components but do not filter masks with a saliency mask.
    %\todo{this table is broken}
    }
    \label{tab:k-ablation}
    \begin{tabular}{lrrrr@{\hskip 12pt}rrrr}
        \toprule
         {$k$} & AP & AP$_\text{S}$ & AP$_\text{M}$ & AP$_\text{L}$ & AR & AR$_\text{S}$ & AR$_\text{M}$ & AR$_\text{L}$\\   %% adjust metrics  
         \midrule
         2 & 0.2 & 0.0 & 0.0 & 0.3 & 0.5 & 0.0 & 0.0 & 2.3 \\%& 0.2 & 0.0 & 0.4 & 0.8 & 3.2 & 0.0 & 0.0 & 13.6 \\
         3 & 2.7 & 0.7 & 1.5 & 8.8 & 4.3 & 0.1 & 1.3 & 16.1 \\%& 0.7 & 0.0 & 0.5 & 2.8 & 5.9 & 0.0 & 0.3 & 24.2 \\
         4 & 3.7 & 0.8 & 2.4 & 12.3 & 6.0 & 0.1 & 2.3 & 21.7 \\%& 1.8 & 0.1 & 1.3 & 4.8 & 7.2 & 0.0 & 1.4 & 28.1 \\
         5 & 3.7 & 0.7 & 3.0 & 11.4 & 6.5 & 0.2 & 3.3 & 22.1 \\%& 1.8 & 0.1 & 1.7 & 4.8 & 8.3 & 0.0 & 2.1 & 31.8 \\
         % Discuss: Why keep these? This sets us up for trouble
         %6 & 0.4 & 0.5 & 0.1 & 1.5 & 4.9 & 0.0 & 1.8 & 17.9 & 1.2 & 0.6 & 2.1 & 3.3 & 7.6 & 0.0 & 3.3 & 27.2\\
         %7 & 0.5 & 0.5 & 0.1 & 1.4 & 4.8 & 0.0 & 2.2 & 16.8 & 1.4 & 0.2 & 2.8 & 3.4 & 7.6 & 0.0 & 4.4 & 25.4\\
         \midrule
         % Discuss: We don't really have these results because we need to filter the masks to train
         % the model in the first place
         2--5 & \textbf{4.5} & \textbf{1.4} & \textbf{6.0} & \textbf{12.8} & \textbf{16.5} & \textbf{1.7} & \textbf{19.1} & \textbf{38.6} \\
         %2--5 & 1.0 & 0.6 & 0.0 & 2.4 & 7.6 & 0.0 & 1.8 & 29.2 & -  \\
        \bottomrule
    \end{tabular}
\end{table}
\begin{table}[h]
  \small
  \centering
  \setlength{\tabcolsep}{4.8pt}
    \caption{\textbf{Ablation of the saliency map.} We also compare different self-supervised models (MAE, DINO) for generating the map. Without the saliency filter (None) performance drops significantly. 
  %The models are evaluated after 30k steps on COCO \texttt{val2017}.
  }
  \label{tab:saliency}
  \begin{tabular}{lcccccc}
  \toprule
  Saliency Mask & AP$_\text{50}$ & AP$_\text{75}$ & AP & AR$_\text{1}$ & AR$_\text{10}$ & AR$_\text{100}$ \\
  \midrule
  None & 2.3 & 0.2 & 0.6 & 0.9 & 3.8 & 9.4 \\

  \midrule
  Ours (MAE~\cite{he2021masked}) & 5.9 & 1.3 & 2.2 & 2.2 & 5.1 & 7.3 \\
  Ours (DINO~\cite{caron2021emerging}) & \textbf{10.7} & \textbf{2.9} & \textbf{4.5} & \textbf{3.5} & \textbf{9.0} & \textbf{16.5} \\
  %Ours (pseudolabels) & & & & & & \\
  \bottomrule
  \end{tabular}
\end{table}
\begin{table}[h]
  \small
  \centering
    \caption{\textbf{Ablation of non-connected component splitting (NCCS)} during the generation of candidate masks. 
  % The models are evaluated after 30k steps on COCO \texttt{val2017}.
  }
  \label{tab:ncc-splitting}
  \begin{tabular}{ccccccc}
  \toprule
  NCCS & AP$_\text{50}$ & AP$_\text{75}$ & AP & AR$_\text{1}$ & AR$_\text{10}$ & AR$_\text{100}$ \\
  \midrule
  \xmark & 7.2 & 2.5 & 3.3 & 3.4 & 6.4 & 9.6 \\
  \cmark & \textbf{10.7} & \textbf{2.9} & \textbf{4.5} & \textbf{3.5} & \textbf{9.0} & \textbf{16.5} \\
  %Ours (pseudolabels) & & & & & & \\
  \bottomrule
  \end{tabular}
\end{table}
\begin{table}[h!]
  \small
  \centering
    \setlength{\tabcolsep}{4.6pt}
      \caption{\textbf{Ablation of the segmentation architecture.} Both networks are trained using MAE mask proposals. 
  % The models are evaluated after 30k steps on COCO \texttt{val2017}.
  }
  \label{tab:arch}
  \begin{tabular}{lcccccc}
  \toprule
  Architecture & AP$_\text{50}$ & AP$_\text{75}$ & AP & AR$_\text{1}$ & AR$_\text{10}$ & AR$_\text{100}$ \\
  \midrule
  Mask~R-CNN~\cite{he2017mask} & 6.8 & 1.8 & 2.7 & 2.2 & 7.3 & 11.9 \\
  SOLOv2~\cite{wang2020solov2, wang2022freesolo} & \textbf{10.7} & \textbf{2.9} & \textbf{4.5} & \textbf{3.5} & \textbf{9.0} & \textbf{16.5} \\
  %Ours (pseudolabels) & & & & & & \\
  \bottomrule
  \end{tabular}
\end{table}

\end{document}